# Subspace Clustering in Wavelet Packets Domain

Ivica Kopriva, *Senior Member, IEEE*, and Damir Seršić, *Member, IEEE*

*Abstract*— Subspace clustering (SC) algorithms utilize the union of subspaces model to cluster data points according to the subspaces from which they are drawn. However, the raw data might not be separable in subspaces, and it should be necessary to find a representation where subspaces are more separable. Furthermore, data points near the intersections of subspaces become source of error when contaminated by noise, and SC algorithms exhibit different sensitivity levels to that. Motivated by these two shortcomings, we propose a wavelet packet (WP) based transform domain subspace clustering. Depending on the number of resolution levels, WP yields several representations instantiated in terms of subbands. The first approach combines original and subband data into one complementary multi-view representation. Afterward, we formulate joint representation learning as a low-rank MERA tensor network approximation problem. That is motivated by the strong representation power of the MERA network to capture complex intra/inter-view dependencies in corresponding self-representation tensor. In the second approach, we use a self-stopping computationally efficient method to select the subband with the smallest clustering error on the validation set. When existing SC algorithms are applied to the chosen subband, their performance is expected to improve. Consequently, both approaches enable the re-use of SC algorithms developed so far. Improved clustering performance is due to the dual nature of subbands as representations and filters, which is essential for noise suppression. We exemplify the proposed WP domain approach to SC on the MERA tensor network and eight other well-known linear SC algorithms using six well-known image datasets representing faces, digits, and objects. Although WP domain-based SC is a linear method, it achieved clustering performance comparable with some best deep SC algorithms and outperformed many other deep SC algorithms by a significant margin. That is in particular case for the WP MERA SC algorithm. On the COIL100 dataset, it achieves an accuracy of 87.45% and outperforms the best deep SC competitor in the amount of 14.75%.

## I. INTRODUCTION

CLUSTERING (a.k.a. unsupervised classification) is one of the fundamental problems in data analysis [1]. Using similarity/distance between data points as a criterion, it aims to infer structure from a set of data points by partitioning (segmenting) them into disjoint homogeneous groups. Many application-specific problems can be formulated as clustering problems, such as image segmentation [2], [3], data mining [4], voice recognition [5], and pattern recognition [6], to name a few. Unfortunately, the high dimensionality of the ambient domain deteriorates clustering performance, and that is related to the well-known phenomenon of the *course of dimensionality*. Hence, the identification of a low-dimensional structure of data in a high-dimensional ambient space is one of the fundamental problems in the fields of engineering and mathematics [7]. By assuming data points lie in a union of subspaces (UoS), they can be clustered by assigning each data point to the subspace from which it is drawn. That stood for motivation to develop algorithms for subspace clustering (SC) [8]-[11]. Unfortunately, the raw data recorded in the ambient domain are not always separable in subspaces. It motivates transform subspace clustering (TSC) [12], with the goal of learning the linear transform in combination with the existing linear SC algorithms such as locally linear manifold clustering [13], low-rank representation SC [8], and sparse SC [9].

The motivation used for TSC stands, in principle, behind the development of nonlinear SC algorithms and tensor-based SC algorithms. The assumption built into the foundation of clustering algorithms is that data within the clusters should have high similarity, while data in different clusters should have low similarity. Related to this is the local invariance assumption [14]. It says that if two data points are close in the original geometry of data distribution, they should stay close in the geometry of their new representation. Thus, cluster labels assigned to data should be invariant to data representation. That is an implicit assumption upon which multi-view SC algorithms are built [15]. Hence, although derivations of kernel-based SC algorithms [16], as well as deep SC algorithms [17]-[25], are often motivated by learning embedding where usage of the UoS model is more justified, these algorithms are also used due to their powerful representation learning abilities. This is especially true for deep SC algorithms. In principle, the equivalent statement applies to tensor models-based SC algorithms. They are focused on learning affinity graphd that capture multi-wise dependencies well, either in single-view SC [26], or intra/inter-view dependencies in multi-view SC [27]-[34].

The presence of noise or errors also limits the performance of SC algorithms. It is intuitively clear that noise will severely affect data points near the intersection of the subspaces, degrading performance of the graph-based methods. Therefore, efforts are made to develop robust SC algorithms [15]. Noise-related performance degradation can be corrected by using a mathematically tractable property known as intrasubspace

Manuscript received ???. This work was supported by the Croatian Science Foundation Grant IP-2022-10-6403. *(Corresponding author: Ivica Kopriva).*

Ivica Kopriva is with the Division of Electronics, Ruđer Bošković Institute, Bijenička cesta 54, 10000 Zagreb, Croatia (e-mail: ikopriva@irb.hr).

Damir Seršić is with the Division of Electronic Systems and Information Processing, Faculty of Electrical Engineering and Computing, University of Zagreb, Unska 3, 10000 Zagreb, Croatia (e-mail: damir.sersic@fer.hr)

projection dominance (IPD) [35]; see Section II.B for more details.

To address the two issues outlined above, we propose a wavelet packets domain (WPD) approach to SC. In comparison with the TSC method, the WPD approach is based on a fast, precomputed discrete wavelet transform. For each data point that is considered to be an image in this paper, WP generates multiple representations instantiated in terms of subbands. Their number depends on the number of resolution levels, and for two resolution levels, twenty representations are generated from one data point. However, subbands can be also understood as outputs of filters. If data are significantly contaminated by noise, the noise suppression aspect of subbands dominates the representation aspect.

In the first approach to WPD SC, we combine original data with four subband representations obtained at the first resolution level (a.k.a. scale) into one five-views data set. As pointed out previously, tensor models-based SC methods [26]-[34] were used extensively to learn good affinity graphs for the later spectral clustering step [36]. Among them are dominat t-product/t-SVD, [37], [38], based methods [29],[30],[33],[34], but the Tucker model [39] and tensor ring model [40], were used as well [31], [32]. However, as emphasized adequately in [27], cited models cannot fully explore the inter- and intra-view information within the self-representation tensor. For that purpose, the multi-scale entanglement renormalization ansatz (MERA) network [41], [42], was proposed in [27]. Its unique architecture, built from isometry-, disentangler- and layer tensors captures naturally inter- and intra-view information. Due to that reason, we apply a low-rank MERA-based algorithm in [27] to our WP-based five-views dataset, achieving outstanding clustering performance, see Section IV.B.

In the second approach to WPD SC, we use a minimum clustering error on the validation subset to select the optimal subband (representation) for a particular data set and a particular SC algorithm. Even though SC algorithms are aimed to operate in a purely unsupervised manner, it is customary to assume in practice that a validation subset exists with labeled samples [44]. In that regard, we use the validation subset for the selection of hyperparameters as well as for the selection of the optimal subband. Afterward, based on the partitions obtained, we estimate orthonormal bases that span individual subspaces. Since the reconstruction of subspace bases from the first $d$ left singular vectors of cluster partitions is equivalent to filtering out the noise associated with small singular values, it is evident why IPD-based postprocessing of the estimated representation matrix improves clustering performance for noise-contaminated data. Clustering out-of-sample (test) data according to subspaces is reduced to finding the minimal distance between a data point and a subspace [44]. The capability to cluster out-of-sample data removes a severe limitation of many existing SC algorithms. As can be seen in Section IV.B, clustering performance on out-of-sample-data closely follows those achieved on in-sample data. We illustrate the proposed WPD approach to SC in Fig. 1.

The main contributions of this paper are as follows.

1) We proposed a WPD approach to multi-view like SC. It combines the original dataset with four subband representations at the first resolution level into a five-views dataset described with a linear multi-view self-representation model. We use a low-rank MERA tensor network for learning a joint affinity graph. We also formulate an out-of-sample extension of the proposed method.

2) We proposed a WPD approach to existing single-view SC algorithms. It transforms the dataset into a number of subbands (representations). Thereby, the subband optimal in terms of a minimum clustering error is selected by a computationally efficient self-stopping rule. We also formulate an out-of-sample extension similar to the previous case for this approach. This approach enables the re-usage of existing SC algorithms with often significantly improved clustering performance.

3) We apply a low-rank MERA and WPD combined with eight linear single-view SC algorithms on six datasets representing digits, faces and objects. As can be seen in Section IV.B, WPD-MERA achieves outstanding clustering performance in all the cases. It outperforms, often with a large margin, even deep SC algorithms. That is achieved despite the fact that WPD-MERA is based on linear WP transform and linear multi-view data model. It can also be seen that linear single-view WPD SC algorithms quite often achieve statistically significant improvement of performance relative to performance in the original domain. In a significant number of cases, the achieved performance is comparable to or even better than the one achieved by deep SC algorithms. Again, that is achieved despite the fact that WPD combined with linear SC algorithms obeys a purely linear data model. MATLAB code of the proposed WP domain SC method is available at https://github.com/ikopriva/WPDSC.

The rest of the paper is organized as follows. In Section II, we review the background and related work. Section III presents our WPD approach to SC. In Section IV, we describe experiments and present results on public datasets. In Section V, we draw conclusions about our study.

## II. BACKGROUND AND RELATED WORK

Table I summarizes the main notation, where transform operators are denoted by calligraphic letters, matrices by bold uppercase letters, vectors by bold lowercase letters, and scalars by italic letters. $\|\mathbf{A}\|_F$ denotes Froebenius norm of the matrix $\mathbf{A}$, $\|\mathbf{A}\|_1$ denotes the $\ell_1$-norm of $\mathbf{A}$, and $\|\mathbf{A}\|_*$ stands for the nuclear norm of $\mathbf{A}$.

### A. Subspace clustering

Let us assume $\mathbf{X} = \{\mathbf{x}_1, \mathbf{x}_2, ..., \mathbf{x}_N\}$ represents a collection of $N$ data points in a $D$-dimensional ambient space. In general, the SC model assumes data points are drawn from $C>1$ affine subspaces of dimensions $\{d_c < D\}_{c=1}^C$, [8]. However, it was shown in [45] that when the dimension of the ambient space is high relative to the sum of the dimensions of affine subspaces, the affine constraint has negligible influence on clustering performance. Thus, in this paper, we shall assume the linear model of the subspaces:

$$\mathbb{S}_c = \left\{ \mathbf{x}_n \in \mathbb{R}^{D \times 1} : \mathbf{x}_n = \mathbf{A}_c \mathbf{z}_n^c \right\}_{n=1}^{N} \quad c \in \{1,...,C\} \ . \quad (1)$$

In the most general sense, the problem of subspace clustering is to identify the number of subspaces $C$, the subspace bases $\{\mathbf{A}_c\}_{c=1}^{C}$, subspace dimensions $\{d_c\}_{c=1}^{C}$, as well as grouping data points according to subspaces from which they are generated [9].

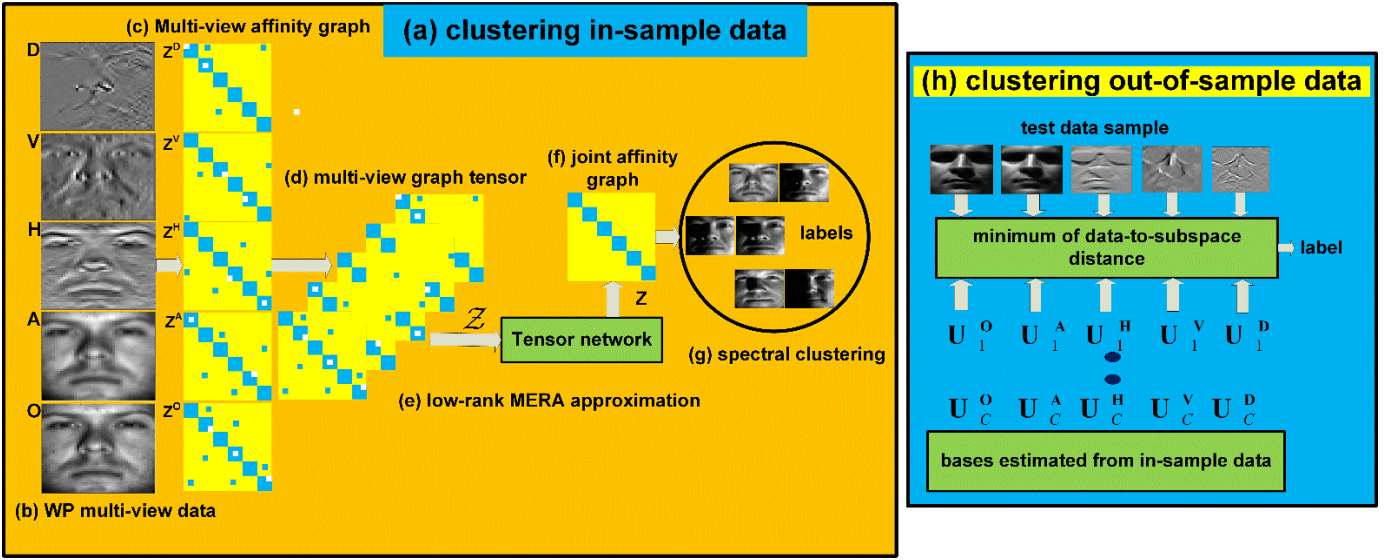

**Fig. 1**. Illustration of WPD-MERA approach to SC on Extended Yale B dataset [64]. (a) Clustering in-sample data. (b) Image data from the original (O) domain, approximation (A) subband, horizontal (H) subband, vertical (V) subband, and details (D) subband are combined into five-views data. (c) Each "view" has its own affinity graph. (d) View-dependent affinity graphs are combined into 3D multi-view graph tensor. (e) A low-rank MERA tensor network is used to estimate (f) joint affinity graph for all views. (g) Spectral clustering is used to cluster in-sample data. (h) Clustering of out-of-sample (test) data. View-dependent bases are estimated from clustered in-sample data. Distances between original and subband representations of test data and corresponding subspaces are calculated. Label corresponding to subspace with minimal distance to data is assigned to test data point. Achieved mean clustering accuracy for a maximal number of clusters on 100 random in-sample partitions for MNIST, USPS, EYaleB, ORL, COIL20, and COIL100 datasets respectively: 99.33%, 99.70%, 99.49%, 88.98%, 99.94%, and 87.45%.

As it is common, we assume that a number of subspaces $C$, which is equivalent to the number of clusters, is known *a priori*. In many SC algorithms, only data clustering, and not identification of subspace bases and subspace dimensions, is required [9], [10], [11]. However, some approaches that follow the UoS model require bases identification. That is necessary for assigning the test (out-of-sample) data points to the subspaces using a criterion based on a point-to-a-subspace-distance [44]; see Section III.E. That is also important for the implementation of the IPD-postprocessing of the representation matrix, see Section II.B.

| Notation | Description |
|---|---|
| $N$ | number of data points |
| $C$ | number of clusters |
| $K$ | total number of nodes in wavelet packets decomposition, i.e. the number of sub-bands across all scales (resolution levels) |
| $N_c$ | number of data points belonging to cluster $c \in \{1,...,C\}$ |
| $\mathbb{S}_c$ | set containing data points belonging to cluster $c \in \{1,...,C\}$ |
| $D$ | dimension of the input (ambient) data space |
| $d_c$ | dimension of the subspace $S_c$ |
| $d$ | assumed equal dimension for all the subspaces $S_c$, $c \in \{1,...,C\}$ |
| $\mathbf{X} \in \mathbb{R}^{D \times N}$ | data matrix comprised of $\left\{ \mathbf{x}_n \in \mathbb{R}^{D \times 1} \right\}_{n=1}^{N}$ data points |
| $\mathbf{E} \in \mathbb{R}^{D \times N}$ | noise (error) matrix comprised of $\left\{ \mathbf{e}_n \in \mathbb{R}^{D \times 1} \right\}_{n=1}^{N}$ points |
| $\mathbf{X}_c \in \mathbb{R}^{D \times N_c}$ | data matrix comprised of data points belonging to cluster $c \in \{1,...,C\}$ |
| $\mathbf{X}^k \in \mathbb{R}^{D \times N}$ | data matrix obtain by wavelet packets transform at node $k \in \{1,...,K\}$ |
| $\mathbf{E}^k \in \mathbb{R}^{D \times N}$ | error matrix obtain by wavelet packets transform at node $k \in \{1,...,K\}$ |
| $\mathbf{X}_c^k \in \mathbb{R}^{D \times N_c}$ | data matrix comprised of data points belonging to cluster $c \in \{1,...,C\}$, at node $k \in \{1,...,K\}$ |
| $k^*$ | index of node with minimal clustering error on $\mathbf{X}^{k^*}$ |
| $\mathbf{A} \in \mathbb{R}^{D \times \sum_{c=1}^{C} d_c}$ | basis of the space spanned by $\mathbf{X}$, $\sum_{c=1}^{C} d_c \leq D$ |
| $\mathbf{Z} \in \mathbb{R}^{\sum_{c=1}^{C} d_c \times N}$ | representation of data $\mathbf{X}$ in basis $\mathbf{A}$ |
| $\mathbf{Z}^k \in \mathbb{R}^{\sum_{c=1}^{C} d_c \times N}$ | representation of data $\mathbf{X}^k$ in basis $\mathbf{A}$ at node $k \in \{1,...,K\}$ |
| $\mathbf{A}_c \in \mathbb{R}^{D \times d_c}$ | basis of the space spanned by $\mathbf{X}_c$, $c \in \{1,...,C\}$ |
| $\mathbf{Z}_c \in \mathbb{R}^{d_c \times N_c}$ | representation of data $\mathbf{X}_c$ in basis $\mathbf{A}_c$, $c \in \{1,...,C\}$ |
| $\mathbf{U}_c \in \mathbb{R}^{D \times d_c}$ | orthonormal basis of the subspace spanned by $\mathbf{X}_c$, $c \in \{1,...,C\}$ |
| $\mathbf{U}_c^{k^*} \in \mathbb{R}^{D_{k^*} \times d_c}$ | orthonormal basis of the subspace spanned by $\mathbf{X}_c^{k^*}$, $c \in \{1,...,C\}$ |
| $\mathcal{W} : \mathbb{R}^D \to \mathbb{R}^{D \times K}$ | Wavelet packets transform applied to data $\{\mathbf{x}\}_{n=1}^N$ |

The data matrix can be represented as: $\mathbf{X} = [\mathbf{X}_1...\mathbf{X}_C]\mathbf{T}$, where $\mathbf{T}$ is an arbitrary permutation matrix. Without loss of generality, we assume $\mathbf{T}=\mathbf{I}$. Based on (1) and assuming the presence of noise and/or errors, we have the following representation of our dataset using a linear subspace model:

$$\mathbf{X} = \mathbf{AZ} + \mathbf{E} \qquad (2)$$

where $\mathbf{A} = [\mathbf{A}_1...\mathbf{A}_C]$, $\mathbf{Z}$ is block diagonal matrix with blocks on the main diagonal $\{\mathbf{Z}_c\}_{c=1}^C$, and $\mathbf{E}$ represents the error term. The term $\mathbf{AZ}$ represents clean but unknown data. Self-expressive model is used in many SC algorithms [8], [9], [10]. It is obtained from (2) by setting $\mathbf{A}=\mathbf{X}$, i.e. each data sample is represented as a linear combination of other data samples. By assuming a normal distribution of the error term, many SC algorithms are obtained as a solution of the optimization problem:

$$\min_{\mathbf{Z}} \tfrac{1}{2}\|\mathbf{X} - \mathbf{XZ}\|_F^2 + \lambda f(\mathbf{Z}) \text{ s.t. } \operatorname{diag}(\mathbf{Z}) = \mathbf{0}. \quad (3)$$

In (3), $f$ is the regularization function imposed on $\mathbf{Z}$, and $\lambda$ stands for the regularization constant. For sparse SC $f(\mathbf{Z}) = \|\mathbf{Z}\|_1$, for low-rank SC [8] $f(\mathbf{Z}) = \|\mathbf{Z}\|_*$, and for locally linear manifold clustering (LLMC) [13] no regularization is imposed on $\mathbf{Z}$. Once $\mathbf{Z}$ is estimated, the data affinity matrix can be obtained as:

$$\mathbf{W} = \frac{|\mathbf{Z}| + |\mathbf{Z}|^\mathsf{T}}{2} \quad . \qquad (4)$$

Once $\mathbf{W}$ is estimated, spectral clustering [36] is applied to a Laplacian matrix to obtain $N \times C$ binary cluster indicator matrix $\mathbf{F} \in Ind$ .

### B. Postprocessing of data affinity matrix

In real-world applications, datasets contain various types of errors. Consequently, data with different labels that lie near the intersections of multiple subspaces are highly likely to be connected with the high-weights edges [35]. That will degrade performance of the graph-based methods such as SC. In [35], a correction method was proposed, and it is based on a mathematically tractable property: intra-subspace projection dominance (IPD) property in the projection (representation) space. IPD says that small coefficients in the representation matrix always correspond to the projections over errors. The effect of errors can be reduced by keeping $d_c$ largest entries and zeroing other entries, where $d_c$ equals to the dimensionality of the corresponding subspace $c \in \{1,...,C\}$. However, to eliminate yet another hyperparameter, we set all subspace dimensions to be equal $\{d_c = d\}_{c=1}^{C}$, and use the existing *a priori* knowledge for $d$, see Section III.E. Thus, we have:

$$\mathbf{Z} \leftarrow \lceil \mathbf{Z} \rceil_d \qquad (5)$$

where the operator $\lceil \mathbf{Z} \rceil_d$ is applied column-wise, keeping $d$ largest coefficients in terms of absolute value and setting others to zero.

### C. Subspace clustering in the transformed domain

It is pointed out in [12] that raw data are not always separable in subspaces, and it is better to transform original data into new representations where they will become more separable, i.e. $\mathcal{T}\mathbf{X} = \mathbf{Y}$. In particular, $\mathcal{T}$ and $\mathbf{Y}$ are learned under the following optimization framework:

$$\min_{\mathcal{T},\mathbf{Y},\mathbf{Z}} \|\mathcal{T}\mathbf{X} - \mathbf{Y}\|_F^2 + \lambda \left( \|\mathcal{T}\|_F^2 - \log \det \mathcal{T} \right) + \mu \|\mathbf{Y}\|_1 \quad (6)$$
$$+ \gamma \|\mathbf{Y} - \mathbf{YZ}\|_2^F + f(\mathbf{Z}).$$

Depending on $f(\mathbf{Z})$, sparse SC, low-rank SC, and LLMC can be optimally tuned to data. Moreover, the generic concept (6) is extended to the kernel version in [12], where subspaces should be more separable. It complements existing kernel SC algorithms such as [16]. Kernel-based SC can be interpreted as representation learning. Analogous conclusion applies to deep SC algorithms.

### D. Wavelets convolutional neural networks

As a computationally efficient alternative to dilated filtering in enlargement of the receptive field of convolutional neural network (CNN), multi-level wavelet CNN was proposed in [49]. It is based on multi-level WP transform [50] that is implemented efficiently through 2D discrete wavelet transform (DWT) [51]. By treating WP transform filters as convolutional filters with predefined weights, WP transform can be seen as a particular type of fully connected network. A wavelet CNN is also proposed in [52] to compensate the property of traditional CNN to miss a large part of spectral information at disposal via multiresolution analysis. In that regard, multiresolution information implemented in terms of filter versions of wavelet function and scaling function is supplemented to traditional CNN. As shown in [52], on texture classification and image annotation tasks, a wavelet CNN outperformed conventional CNN while having significantly fewer parameters. In order to provide a mathematical understanding of deep convolutional networks' ability to build large-scale invariants stable to deformations, an invariant scattering convolution network was proposed in [53], [54]. The scattering transform network implements a cascade of filters that compute the wavelet transform and a pointwise nonlinearity. Wavelet transform is necessary to separate variations of data instances, and nonlinearity is necessary to preserve invariance to translation. In terms of architecture, scattering transform looks very much like a WP transform. In our approach to WPD SC, we do not use nonlinearities because we want to preserve information on subspaces, see the next section.

### E. Independent subspace analysis

WP transform was also used to solve the blind source separation (BSS) problem comprised of statistically dependent sources [55], [56]. There, original univariate mixture data were transformed in the WP domain, where a subband with the least statistically dependent components was selected. Afterward, independent component analysis (ICA) was applied to the selected subband to recover the mixing matrix of the original BSS problem. That concept is extendable to independent subspace analysis (ISA) [57]. To see that, we emphasize that the **AZ** term in (2) represents the multidimensional ICA (MICA) model [58]. In the MICA model, sources are random vectors (multidimensional random variables) that are mutually statistically independent, but one-dimensional random variables within each random vector can be statistically dependent. The same reasoning applies to data matrix **X** when data within the same cluster are treated as random vectors. ISA problem is concerned with the identification of subspaces up to the ambiguities related to permutation of subspaces and invertible linear transformations within the subspaces [59]. In our approach to WPD SC we are not interested in the subspace identification problem, but only in the SC problem, i.e. assigning data points according to the subspaces they are generated from.

### F. Tensor subspace clustering

Tensor models-based SC methods [26]-[34], were used extensively to learn good affinity graphs for the later spectral clustering step. Among them dominate t-product/t-SVD, [37], [38], based methods [29], [30], [33], [34]. It is, however, properly emphasized in [27] that t-SVD models, as well as Tucker and tensor ring models, cannot fully explore the inter- and intra-view information within the self-representation tensor. Therefore, the multi-scale entanglement renormalization ansatz (MERA) network [41], [42] was proposed in [27]. In multi-view SC, low-rank MERA aims to approximate self-representation tensor $\mathcal{Z} \in \mathbb{R}^{N \times N \times V}$ based on the MERA tensor network. For that purpose, $\mathcal{Z}$ is reshaped into 5D tensor $\mathcal{Y} \in \mathbb{R}^{I_1 \times I_2 \times I_3 \times I_4 \times I_5}$, where $I_1=A$, $I_2=Q$, $I_3=A$, $I_4=Q$, $I_5=V$, and $N=A \times Q$. Low-rank MERA approximation of $\mathcal{Y}$ is to find MERA factors: isometries $\mathcal{W}_1^i \in \mathbb{R}^{I_1 \times I_2 \times R_1}$ and $\mathcal{W}_2^i \in \mathbb{R}^{I_3 \times I_4 \times I_5 \times R_2}$, disentangler $\mathcal{U}_1 \in \mathbb{R}^{I_2 \times I_3 \times I_2 \times I_3}$, and top core $\mathbf{B} \in \mathbb{R}^{R_1 \times R_2}$, i.e. $f\left(\mathbf{B}, \mathcal{W}_1^i, \mathcal{W}_2^i, \mathcal{U}_1\right)$. The related optimization problem is formulated as [27]:

$$\min_{\mathcal{U}_1, \mathcal{W}_1^i, \mathcal{W}_2^i, \mathbf{B}} \frac{1}{2}\left\|\mathcal{Y} - f\left(\mathbf{B}, \mathcal{W}_1^i, \mathcal{W}_2^i, \mathcal{U}_1\right)\right\|_F^2 \qquad (7)$$

where it is assumed $R_1=R_2=R$. MERA-based multi-view SC problem is formulated as [27]:

$$\min_{\left\{\mathbf{Z}^{(v)}, \mathbf{E}^{(v)}\right\}_{v=1}^{V}} \sum_{v=1}^{V} \lambda \|\mathbf{E}\|_{2,1}$$
$$\text{s.t.} \mathbf{X}^{(v)} = \mathbf{X}^{(v)} \mathbf{Z}^{(v)} + \mathbf{E}^{(v)} \ v=1,...,V \qquad (8)$$
$$\mathcal{Y} = f\left(\mathbf{B}, \mathcal{W}_1^i, \mathcal{W}_2^i, \mathcal{U}_1\right)$$

Thus, the low-rank MERA has two hyperparameters: $\lambda$ and $R$. After the optimization is finished, we compute the layer tensor $\mathcal{C} \in \mathbb{R}^{I_1 \times I_2 \times I_3 \times I_4 \times I_5 \times R \times R}$ as:

$$\mathcal{C} = \mathcal{U}_1 \times_{\{I_2\}} \mathcal{W}_1^i \times_{\{I_3\}} \mathcal{W}_2^i. \qquad (9a)$$

We compute the top core as:

$$\mathbf{B} = \mathcal{Y} \times_{\{I_1, I_2, I_3, I_4, I_5\}} \mathcal{C}. \qquad (9b)$$

We compute estimate of $\mathcal{Y}$ as:

$$\hat{\mathcal{Y}} = \mathcal{C} \times_{\{R_1, R_2\}} \mathbf{B} \qquad (9c)$$

and estimate of $\mathcal{Z}$ is obtained as :

$$\hat{\mathcal{Z}} \in \mathbb{R}^{N \times N \times V} = reshape\left(\hat{\mathcal{Y}}\right). \qquad (9d)$$

In (9a) to (9c) $\times_{\{I_n\}}$ denotes contraction over mode $n$. A unified representation for all the views is obtained from $\hat{\mathcal{Z}}$ through averaging over mode 3 (the view mode). Unique architecture of the MERA network enables capturing inter- and intra-view information. Therefore, we apply the low-rank MERA-based algorithm to our WP-based five-views dataset, achieving outstanding clustering performance.

## III. METHOD

This section details our approach to WPD SC. In this work, we limit ourselves to datasets comprised of vectorized images. Thus, we apply 2D DWT with four wavelet filters on each metricized version (image) $\left\{ \mathbf{X}_n \right\}_{n=1}^{N}$ of data points $\left\{ \mathbf{x}_n \right\}_{n=1}^{N}$. The wavelet filters are low-pass filter $h_{LL}$, band-pass filters $h_{LH}$ and $h_{HL}$, and high-pass filter $h_{HH}$. To simplify notation we adopt the customary notational convention of approximation A↔LL, horizontal H↔LH, vertical V↔HL, and details D↔HH. That implements the decomposition at the first resolution level, and we call filtered data subbands. Each subband can be further decomposed into four subbands, yielding sixteen subbands at the second resolution level. The process can continue recursively, yielding a quaternary decomposition tree. The tree nodes correspond to the subbands at the appropriate scale (resolution level). In implementing the WPD approach to SC, we use the Haar wavelet.

### A. Wavelet packets transform for subspace clustering

By denoting the total number of subbands with $K$ we obtain the following:

$$\mathcal{W}\left(\mathbf{X}_n\right) \rightarrow \left\{\mathbf{X}_n^k + \mathbf{E}_n^k\right\}_{k=1}^{K} \quad n \in \{1, ..., N\}. \tag{10}$$

The index $k$ in (10) represents a combination of the sub-band index and scale index. Representation of each data point at the decomposition level $k$ is expressed in terms of its decomposition coefficients [61]:

$$z_n^k\left(\xi\right) = \sum_l h_{nl}^k \varphi_l\left(\xi\right) \tag{11}$$

where $l$ represents the shift index, and $\xi$ is an independent variable that, in the case of subspace clustering, corresponds to the feature index. Each data point at decomposition level $k$ is expressed in terms of its decomposition coefficients [60]:

$$x_n^k\left(\xi\right) = \sum_l f_{nl}^k \varphi_l\left(\xi\right) . \tag{12}$$

We also expand the error term on the same wavelet basis:

$$e_n^k\left(\xi\right) = \sum_l g_{nl}^k \varphi_l\left(\xi\right) . \tag{13}$$

By using the orthogonality property of $\varphi_l\left(\xi\right)$ it follows [60]:

$$\mathbf{f}_l = \mathbf{A}\mathbf{h}_l + \mathbf{g}_l \tag{14}$$

Inserting (14) into (12) and using (11) and (13), we obtain:

$$\mathbf{x}_n^k = \mathbf{A}\mathbf{z}_n^k + \mathbf{e}_n^k \quad n \in \{, ..., N\}1; k \in \{1, ..., K\}. \tag{15}$$

or on the matrix level:

$$\mathbf{X}^k = \mathbf{A}\mathbf{Z}^k + \mathbf{E}^k \quad k \in \{1, ..., K\}. \tag{16}$$

Direct comparison between (2) and (16) implies that WP-transformed data follow the same UoS model as data in the original ambient space. In other words, the subspace information contained in bases $\mathbf{A}$ is preserved. That is why we do not apply pointwise nonlinearities as in the scattering transform [53], [54]. It can be seen in Table I that dimensions of WP transformed data are assumed to be equal to dimension $D$ of the original ambient space. That is because we use a non-decimated implementation of 2D DWT. The most important differences between (2) and (16) are: (*i*) WP transform generates $K$ representations where some of them should separate data from different groups better than in (2); (*ii*) the error term in (13) is subband dependent, i.e., it is the result of the filtering of the original error term in (2). If the data set is contaminated by noise or if the chosen SC algorithm is sensitive to the presence of noise, subbands that suppress noise, such as A or AA, will be preferred. Instead, subbands that ensure increased separability between data points belonging to different subspaces will be preferred for other combinations of datasets and SC algorithms. Thus, the proposed WPD approach to SC offers adaptability to datasets and SC algorithms.

### B. Wavelet packets and MERA network for subspace clustering

Herein, we propose to apply the low-rank MERA approximation network described in Section II.F to multi-view like SC. For that purpose, we combine original data, denoted herein as $\mathbf{X}^O$, with four representations obtained by WP transform at the first resolution level, namely $\mathbf{X}^A$, $\mathbf{X}^H$, $\mathbf{X}^V$ and $\mathbf{X}^D$, into the five-views data set $\left\{\mathbf{X}^v \in \mathbb{R}^{D \times N}\right\}_{v \in \{O, A, H, V, D\}}$. We use the MERA multi-view SC algorithm to estimate the self-representation tensor $\hat{\mathcal{Z}} \in \mathbb{R}^{N \times N \times 5}$. The self-representation matrix that unifies all the views is obtained as follows:

$$\hat{\mathbf{Z}} = \frac{1}{5}\sum_{v=1}^{5} \hat{\mathcal{Z}}\left(:,:,v\right) . \tag{17}$$

The spectral clustering algorithm [36] is now applied to $\hat{\mathbf{Z}}$ in order to assign labels to data points, i.e. to partition data into $C$ clusters:

$$\bigcup_{c=1}^{C} \mathbf{X}_c^v = \mathbf{X}^v, v \in \{O, A, H, V, D\} . \tag{18}$$

We describe in Section III.E how the proposed method is applied to cluster out-of-sample (test) data.

### C. Subspace clustering on best subband

The proposed approach to WPD SC can be used with existing linear single-view SC algorithms to improve their performance. For that purpose, we need to select optimal subband for particular SC algorithm and data set. In that regard, we propose to use a minimum of the clustering error (*CE*) criterion on a validation subset, i.e.:

$$k^* = \arg\min_{k \in \{1, ..., K\}} CE\left(\mathbf{X}^k\right) \tag{19}$$

When computing $CE$ in (19), we assume that hyperparameters of the specific SC algorithms are also tuned on selected validation subset. A naive implementation of the proposed approach requires evaluation of (19) over $K$ subbands. For two or three resolution levels that respectively implies 20 and 88 subbands. However, we compare the smallest $CE(k^1)$ on the first resolution level with the $CE(0)$ of the original data. In case $CE(0) < CE(k^1)$, we accept original data $\mathbf{X}$ as optimal and stop the process. Otherwise, if a number of resolution levels is greater than one, we apply WP transform to $\mathbf{X}^{k^1}$. Evidently, $k^1 \in \{A, H, V, D\}$. We now estimate $CE(k^2)$, where $k^2 \in \{k^1A, k^1H, k^1V, k^1D\}$. If $CE(k^1) < CE(k^2)$ we accept $\mathbf{X}^{k^1}$ as optimal and stop. Otherwise, if number of resolution levels is two, we accept $\mathbf{X}^{k^2}$ as optimal and stop. On the opposite, we apply WP transform to $\mathbf{X}^{k^2}$ and repeat the procedure. In our experimental setting, we worked with two resolution levels. In that case, the described approach needs evaluation of $CE$ over eight subbands plus original data, while naive implementation requires twenty subbands plus original data. We summarize the proposed WPD SC of the in-sample data in Algorithm 1.

---

**Algorithm 1** Subspace clustering in WP domain

**Inputs**: In-sample dataset $\mathbf{X} = \{\mathbf{x}_n\}_{n=1}^N$, $C$ - number of clusters, $J$ number of resolution levels.

**Initialize**: Current resolution level: $j=1$; best subband from previous resolution level: $k^0=0$.

*Step 1*: Apply WP transform column-wise to $\mathbf{X}$ according to (11).

*Step 2*: Estimate $CE$ for the original data, $CE(k^0)$.

*Step 3*: Estimate clustering error ($CE$) on each subband $\{\mathbf{x}_n^k\}_{n=1}^N$, $k \in \{k^0A, k^0H, k^0V, k^0D\}$ to detect sub-band $k^1$ with the smallest $CE$.

*Step 4*: **If** $CE(k^0) < CE(k^1)$: $\mathbf{X}^{k^*} = \mathbf{X}$. Go to *Step 6*.

*Step 5*: **If** $CE(k^1) < CE(k^0)$ and $j<J$:

$\quad$ $j=j+1$; $k^0=k^1$;

$\quad\quad$ Go to *Step 3*.

$\quad$ **else**

$\quad$ $\mathbf{X}^{k^*} = \mathbf{X}^{k^1}$. Go to *Step 6*.

$\quad$ **end**

*Step 6*: Apply the chosen SC algorithm on $\mathbf{X}^{k^*}$.

**Output**: $k^*$ - optimal subband index; partitions $\{\mathbf{X}_c^{k^*}\}_{c=1}^C$, such that $\bigcup_{c=1}^C \mathbf{X}_c^{k^*} = \mathbf{X}^{k^*}$; binary cluster assignment matrix $\mathbf{F} \in \mathbb{N}_{0+}^{N \times C}$.

---

### D. Geometric interpretation of performance improvement

It is discussed in Section III.C how WPD SC is expected to adapt to the specific combination of SC algorithm and dataset. If representation quality is influenced dominantly by noise suppression, we expect low-pass subbands such as A or AA to be preferred. If the quality of representation is influenced dominantly by the separation of data points belonging to different subspaces, band-pass or high-pass subbands, such as AH, D, or DH, are expected to be preferred. We can verify these hypotheses by estimating distances between subspaces spanned by obtained partitions. Compared with the clustering in the original input space, subspaces in the former case should be closer, while in the latter case they should be far away from each other. To that end, let $\phi_1 \leq \phi_2 \leq ... \leq \phi_d$ be $d$ principal angles between the two $d$-dimensional subspaces $\mathbb{S}_1$ and $\mathbb{S}_2$. Let $\mathbf{U}_1 \in \mathbb{R}^{D \times d}$ and $\mathbf{U}_2 \in \mathbb{R}^{D \times d}$ be orthonormal bases for $\mathbb{S}_1$ and $\mathbb{S}_2$ respectively. Affinity, as a measure of similarity, between the two subspaces can be calculated as [61]:

$$\left\| \mathbf{U}_1^\mathsf{T} \mathbf{U}_2 \right\|_\sigma = \sqrt{\frac{\sum_{i=1}^d \cos^2 \phi_i}{\hat{d}}} \tag{20}$$

where $\hat{d} = \min(d_1, d_2)$. In (20), $\left\| \mathbf{U}_1^\mathsf{T} \mathbf{U}_2 \right\|_\sigma = 1$ if $\mathbb{S}_1 = \mathbb{S}_2$, and $\left\| \mathbf{U}_1^\mathsf{T} \mathbf{U}_2 \right\|_\sigma = 0$ if $\mathbb{S}_1 \perp \mathbb{S}_2$. We define the average affinity between subspaces in the ambient input space as:

$$affinity(\mathbf{X}) = \frac{2}{C \times (C-1)} \sum_{i=1}^{C-1} \sum_{j=i+1}^{C} \left\| \mathbf{U}_i^\mathsf{T} \mathbf{U}_j \right\|_\sigma. \tag{21}$$

The average affinity between subspaces in the WP-space is defined as:

$$affinity(\mathbf{X}^{k^*}) = \frac{2}{C \times (C-1)} \sum_{i=1}^{C-1} \sum_{j=i+1}^{C} \left\| (\mathbf{U}_i^{k^*})^\mathsf{T} \mathbf{U}_j^{k^*} \right\|_\sigma \tag{22}$$

For subband $k^*$ that increases the separation between data points belonging to different subspaces, we expect:

$$affinity(\mathbf{X}^{k^*}) < affinity(\mathbf{X}) \tag{23}$$

For subband $k^*$ that filters out noise, we expect the opposite of (23).

### E. Clustering out-of-sample data

Many SC algorithms are incapable of clustering out-of-sample (a.k.a. unseen or test) data [8]-[13], [16]. That also applies to deep SC algorithms [18]-[25] and tensor-based SC algorithms [26]-[34]. In other words, to cluster the unseen data point, the algorithm has to be re-run again on a dataset enlarged with the unseen data point. That hinders the applicability of these algorithms to large-scale- and/or online clustering problems. Herein, we formulate the problem of clustering the out-of-sample data point as a minimization of the point-to-a-subspace distance criterion.

*1) Wavelet packets MERA subspace clustering*

We use partitions (18) obtained by the WP MERA SC algorithm on the in-sample dataset to estimate the subspace bases [44]:

$$\left\{ \mathbf{X}_c^v \leftarrow \mathbf{X}_c^v - \left[ \underbrace{\overline{\mathbf{x}}_c^v \dots \overline{\mathbf{x}}_c^v}_{N_c \text{ times}} \right] \right\}_{c=1}^C, \ v \in \{ \text{O,A,H,V,D} \} \ (24)$$

where $\overline{\mathbf{x}}_c^v = \frac{1}{N_c} \sum_{n=1}^{N_c} \mathbf{X}_c^v(n)$ , $\bigcup_{c=1}^C \mathbf{X}_c^v = \mathbf{X}^v$ , and $\sum_{c=1}^C N_c = N$ .

From $\left\{ \mathbf{X}_c^v = \mathbf{U}_c^v \Sigma_c^v \left( \mathbf{V}_c^v \right)^T \right\}_{c=1}^C$ we estimate orthonormal bases from the first $d$ left singular vectors of partitions, i.e. $\left\{ \mathbf{U}_c^v \in \mathbb{R}^{D \times d} \right\}_{c=1}^C$ [44]. For test data point $\mathbf{x}^v$ we measure point-to-subspace distances $\left\{ d_{c_v}^v = \left\| \tilde{\mathbf{x}}_c^v - \mathbf{U}_c^v \left( \mathbf{U}_c^v \right)^T \tilde{\mathbf{x}}_c^v \right\|_2 \right\}$ :

$$c_v = \arg\min_{c \in \{1,\dots,C\}} d_{c_v}^v, \ v \in \{ \text{O,A,H,V,D} \} \quad (25)$$

where $\tilde{\mathbf{x}}_c^v = \mathbf{x}^v - \overline{\mathbf{x}}_c^v$ . We assign label $\{c\}_{c=1}^C$ to test data point:

$$\left[ \pi(\mathbf{x}) \right]_c = \begin{cases} 1, \text{ if } c = \arg\min_{c_v \in \{c_O, c_A, c_H, c_V, c_D\}} d_{c_v} \\ 0, \text{ otherwise.} \end{cases} \quad (26)$$

*2) Subspace clustering on the best subband*

For clustering test data on best sub-band $k^*$, we use partitions obtained by Algorithm 1. We estimate orthonormal bases $\left\{ \mathbf{U}_c^{k^*} \in \mathbb{R}^{D \times d} \right\}_{c=1}^C$ with a procedure analogous to the one presented above. We assign a label $\{c\}_{c=1}^C$ to the test point in WP domain, $\mathbf{x}^{k^*}$, according to the point-to-a-subspace distance criterion :

$$\left[ \pi\left( \mathbf{x}^{k^*} \right) \right]_c = \begin{cases} 1, \text{ if } c = \arg\min_{l \in \{1,\dots,C\}} \left\| \tilde{\mathbf{x}}^l - \mathbf{U}_l^{k^*} \left( \mathbf{U}_l^{k^*} \right)^T \tilde{\mathbf{x}}^l \right\|_2 \\ 0, \text{ otherwise.} \end{cases} \quad (27)$$

where $\tilde{\mathbf{x}}^l = \mathbf{x}^{k^*} - \overline{\mathbf{x}}_l^{k^*}$ .

Subspace dimension $d$ is hyperparameter. However, in many scenarios it is known. For example, face images of each subject in the Yale B dataset lie approximately in a $d$=9 subspace [46]. Handwritten digits, lie approximately in a $d$=12 subspace [47]. Regarding the COIL-20 and COIL-100 datasets, the recommended subspace dimensions are $d$=9, [48].

## IV. EXPERIMENTS AND RESULTS

This section evaluates the proposed WPD SC on six benchmark datasets representing digits, faces and objects. We compare the performance of eight well-known linear SC algorithms in the original ambient domain, and the best subband domain. Our intention was to validate relative performance improvement due to clustering in the WP domain. Furthermore, we also validate the clustering performance of the WP MERA SC algorithm. For each dataset, we cite the reported performance of several deep SC networks in order to emphasize the quality of clustering results achieved by linear SC algorithms in the WP domain.

### A. Experimental setup

*Software environment.* All experiments were performed in MATLAB 2021a software environment on a computer with 256 GB of RAM, with a 2.2 GHz Intel Xeon CPU E5-2650 v4 2 processors.

*Evaluation metrics.* Following the convention in the clustering literature, for example [27], we use five metrics for comparative performance analysis in reported experiments: accuracy (ACC), normalized mutual information (NMI), Rand index, F-score and purity. All metrics belong to [0, 1] interval with 0 meaning the worst-, and 1 meaning the best performance.

*Benchmark datasets.* We use six benchmark datasets to evaluate the performance of proposed WP SC algorithms: MNIST [62], USPS [63], EYaleB [64], ORL [65], COIL20 and COIL100 [66]. Table II shows the main characteristics of these datasets. MNIST and USPS contain digit images, ORL and EYaleB contain face images, and COIL20 and COIL100 contain images of objects.

TABLE II
MAIN CHARACTERISTICS OF DATASETS USED IN THE EXPERIMENTS

| Dataset | #Sample | #Feature | #Cluster |
|---------|---------|----------|----------|
| MNIST | 10000 | 28×28 | 10 |
| USPS | 7291 | 16×16 | 10 |
| EYaleB | 2432 | 48×42 | 38 |
| ORL | 400 | 32×32 | 40 |
| COIL20 | 1440 | 32×32 | 20 |
| COIL100 | 7200 | 32×32 | 100 |

*Compared methods.* First, we validate the performance of the WP MERA SC algorithm. Although tensor-based methods [26]-[34] yield high-quality results in SC, we do not report their performance herein due to two reasons: (*i*) it is shown in [27] that low-rank MERA multi-view SC outperformed them; (*ii*) WP MERA SC algorithm yielded virtually perfect clustering performance, see Tables IV to IX. That is achieved even though "views" were comprised of original data and four subbands at the first resolution level. In contrast, in [27] various feature constructors such as Gabor, LBP, HOG, GIST, etc., were used for this purpose. WP MERA SC algorithm has two hyperparameters $\lambda$ and $R$, see Section II.F. We selected them by grid search such that $\lambda \in \{10^{-10}, 10^{-9}, 10^{-8}, 10^{-7}, 10^{-6}, 10^{-5}, 10^{-4}, 10^{-3}, 10^{-2}, 10^{-1}\}$ and $R \in \{2:1:20\}$. Selected values for each dataset are reported in Table III. We validated eight linear single-view SC algorithms in the original domain and the best subband domain: (1) sparse SC (SSC) [9]. The SSC algorithm can be run in two modes: assuming additive white Gaussian noise and assuming outliers. That is why possible performance improvement of the SSC algorithm is important. When tuning the algorithm in ambient domain and WP domain, we varied the parameter $\alpha \in \{1:1:40\}$; (2) Generalization of minimax concave penalty low-rank sparse SC (GMC-LRSSC) [10]. The algorithm has three hyperparameters: low-rank ($\lambda$) and sparsity ($1-\lambda$) constraint which are additionally parameterized with $\alpha/(1+\lambda)$, and (non)-convexity parameter $\gamma$. $\lambda$ was tuned in the range 0 to 1 with the step 0.1. $\alpha$ was tuned in the range $\alpha \in \{10^{-3}, 10^{-2}, 10^{-1}, 1, 10:1:20, 25:5:60, 10^2, 10^3\}$. (Non)-convexity parameter was tuned in the range $\gamma \in \{0.0:0.1:1\}$, 1}; (3) $S_0$-$\ell_0$ low-rank sparse SC (S0L0-LRSSC) [10]. The algorithm is parameterized in terms of $\lambda$ and $\alpha$ in a way analogous to the GMC-LRSSC. Due to space limitations, for each dataset we only report better results between GMC- and S0L0 LRSSC algorithms in the main paper. Complete results are reported in the supplement; (4) Low-rank representation (LRR) SC [8]. LRR learns the low-rank representation matrix in the self-representation data model, where the low-rank constraint is implemented in terms of the $S_1$ norm. The algorithm is parameterized in terms of $\lambda \in \{0.1:0.1:0.9\}$.; (5) Nearest subspace neighbor (NSN) algorithm [67] determines first the neighborhood set of each data point and, afterward, uses greedy subspace recovery algorithm to estimate subspace from the given set of points. The algorithm has two hyperparameters: the number of nearest neighbors, $k$, and the maximal subspace dimension $d_{max}$. We estimated $k$ from the set $k \in \{2, 3, 4, 8, 10, 12, 15, 20, 25, 30, 35, 37, 40, 45\}$. Maximal dimension was set either to $d_{max}=k$ or to $d_{max}=d$, where $d$ is subspace dimension; (6) The robust thresholding SC (RTSC) algorithm [11], estimates the adjacency matrix by estimating the set of $q$ nearest neighbors for each data point according to the metric $s(\mathbf{x}_i, \mathbf{x}_j) := \arccos(|\langle \mathbf{x}_i, \mathbf{x}_j \rangle|)$. The parameter $q$ is estimated according to $q = \max(k, \lceil N_c/20 \rceil)$. Essentially, this step is equivalent to the IPD step (5). We selected $k$ from the set $k \in \{2:20\}$; (7)/(8) $S_{1/2}$-LRR and $S_{2/3}$-LRR SC algorithms [68]. These are low-rank regularized SC methods, where low-rank constraints are implemented in terms of Schatten $S_{1/2}$ and $S_{2/3}$ norms. The algorithms have one regularization constant selected from the set $\lambda \in \{0.01, 0.05:0.05:1, 1.5, 2:10\}$. Due to space limitation, we only report the best result between $S_1$, $S_{1/2}$ and $S_{2/3}$ norm-constrained LRR algorithms. In terms of preprocessing, all data were column-normalized. We present hyperparameters tuned for each SC algorithm and each dataset in Table III.

TABLE III
TUNED VALUES OF THE REGULARIZATION CONSTANTS FOR CHOSEN SC ALGORITHMS AND CHOSEN DATASETS IN THE AMBIENT DOMAIN (LEFT) AND THE WP DOMAIN (RIGHT).

| | MNIST | USPS | EYaleB | ORL | COIL20 | COIL100 |
|---|---|---|---|---|---|---|
| WP MERA SC | $\lambda=10^{-4}$ R=3 | $\lambda=10^{-2}$ $R=6$ | $\lambda=10^{-3}$ $R=10$ | $\lambda=10^{-9}$ $R=10$ | $\lambda=0.1$ $R=13$ | $\lambda=0.1$ $R=17$ |
| SSC | $\alpha=(6, 5)$ outl. outl. affine | $\alpha=(3, 3)$ | $\alpha=(15, 10)$ outl. outl. | $\alpha=(19, 14)$ affine outl. affine | $\alpha=(7, 12)$ | $\alpha=(20, 4)$ |
| GMC-LRSSC | $\lambda=(0.1, 0.1)$ $\alpha=(46, 50)$ $\gamma=(0.7, 1)$ | $\lambda=(0.1, 0.1)$ $\alpha=(13, 11)$ $\gamma=(0.8, 0.9)$ | $\lambda=(29, 0.1)$ $\alpha=(2.5, 8)$ $\gamma=(1, 0.6)$ | $\lambda=(1.4, 2.3)$ $\alpha=(0.8, 1)$ $\gamma=(0.1, 0.5)$ | $\lambda=(10, 0.1)$ $\alpha=(6, 61)$ $\gamma=(0.7, 0.5)$ | $\lambda=(9.5, 0.1)$ $\alpha=(6, 26)$ $\gamma=(0, 0.9)$ |
| S0L0-LRSSC | $\lambda=(0.3, 0.35)$ $\alpha=(24, 9)$ | $\lambda=(0.35, 0.35)$ $\alpha=(9, 8)$ | $\lambda=(0, 0.5)$ $\alpha=(3, 9)$ | $\lambda=(0.3, 0.4)$ $\alpha=(3, 12)$ | $\lambda=(0, 0.5)$ $\alpha=(19, 24)$ | $\lambda=(0.8, 0.4)$ $\alpha=(20, 22)$ |
| LRR | $\lambda=(0.4, 0.3)$ | $\lambda=(0.2, 0.2)$ | $\lambda=(2, 0.7)$ | $\lambda=(1.5, 1.85)$ | $\lambda=(0.4, 0.1)$ | $\lambda=(0.1, 0.6)$ |
| NSN | $k=(36, 31)$ $d_{max}=(12, 12)$ | $k=(60, 54)$ $d_{max}=(11, 11)$ | $k=(32, 42)$ $d_{max}=(k, k)$ | $k=(6, 3)$ $d_{max}=(9, 8)$ | $k=(24, 17)$ $d_{max}=(k, k)$ | $k=(42, 31)$ $d_{max}=(9, k)$ |
| RTSC | $q=(18, 6)$ | $q=(23, 23)$ | $q=(2, 7)$ | $q=(3, 4)$ | $q=(4, 5)$ | $q=(2, 4)$ |
| $S_{1/2}$-LRR | $\lambda=(0.25, 0.15)$ | $\lambda=(0.16, 0.16)$ | $\lambda=(6, 0.6)$ | $\lambda=(7.5, 20)$ | $\lambda=(0.6, 0.1)$ | $\lambda=(0.3, 0.01)$ |
| $S_{2/3}$-LRR | $\lambda=(0.15, 0.2)$ | $\lambda=(0.15, 0.15)$ | $\lambda=(4, 0.8)$ | $\lambda=(4.5, 0)$ | $\lambda=(0.9, 0.1)$ | $\lambda=(0.15, 0.01)$ |

We tuned hyperparameters on ten randomly selected subsets using average accuracy as a criterion. A number of samples per group was for MNIST, USPS, EYaleB, ORL, COIL20, and COIL100 in respective order: 50, 50, 43, 7, 50, and 50. The motivation for using linear single-view SC algorithms in the WP domain was to verify whether they can be re-used with improved clustering performance. As can be seen, in many cases, performance is statistically significantly improved in comparison with performance in the original ambient domain.

To obtain good assessment of the quality of clustering performance achieved by proposed WPD SC algorithms, we also reported performance of several deep SC algorithms in cited references: deep structure learning with similarity preserving (DSLSP) [69], adaptive attribute and structure subspace clustering network (AASSC-Net) [19], deep SC network (DSC) [17], deep adversarial SC (DASC) [21], pseudo-supervised deep SC (PSSC) [22], maximum entropy SC (MESC) [18], structured auto-encoder (SAE) for SC [23], deep cognitive SC (DCSC) [24], AE-based latent block-diagonal representation for SC (LBDR) [25], low-rank constrained autoencoder (LRAE) [70], deep subspace image clustering network with self-representation and self-supervision (DSCNSS) [71], deep self-representation subspace clustering network (DSRSCN) [72], and deep closed-form subspace clustering [73]. In Tables VI and VIII, we also cited results from [12] related to learnable transformed domain SSC, LR SC, and LLMC on EYaleB and COIL20 datasets.

*B. Clustering performance on benchmark datasets*

Tables IV to IX present results of comparison of the proposed WPD SC algorithms with state-of-the-art deep SC algorithms [17], [18], [19], [21]-[25], [69], [7] and transformed domain SC algorithms [12]. In the case of WPD SC algorithms, we also applied IPD-based postprocessing of the estimated representation matrix. We reported these results if they were better than those obtained by the WPD SC algorithms. The best result is in bold font. The second-best result is underlined. It is seen that the performance of the WP MERA SC algorithm is the best for all datasets, with the exception of the ORL dataset, where it is the second best. However, since WP MERA was validated on 100 randomly selected subsets, we argue that it actually outperforms the cited deep SC algorithm. The most striking difference is in the case of the COIL100 dataset, where WP MERA SC outperforms the best deep SC algorithm by 14.75% in accuracy. In the case of MNIST, WP MERA outperforms the best deep SC by 7.69% in accuracy, and by 16.36% in NMI, while in the case of USPS the improvements in respective orders are 16.41% and 15.69%. In the case of EYaleB, improvements are 0.49% and 5.89 in comparison with the TSC SSC [12]. Note, however, that TSC SSC is a learnable approach, while the WP approach is based on pre-computed filter bank-based DWT. Also, note that TSC LRR yields poor results. In the case of the ORL dataset, the performance of WP MERA, AASSC, and MESC networks are virtually the same, and it is much better than the performance of other reported deep SC algorithms. In the case of the COIL20 dataset, WP MERA outperforms TSC LLMC [15] by 1.94% in accuracy and by 5.93% in NMI. It also outperforms the MESC network by 1.54% in accuracy and by 1.64% in NMI. Reported results are outstanding, taking into account that, as opposed to deep SC algorithms, WP MERA SC is based on the linear data model. We also applied the Wilcoxon test to verify whether SC in WPD on the best sub-band yielded statistically significantly improved performance. Cases with the *p*-values greater than 0.05 are shaded in grey. Except for the USPS dataset, SC in WPD yields significantly improved performance in the case of the majority of considered linear single-view SC algorithms. That is true for both in-sample and out-of-sample data.

Regarding the performance of individual linear single-view SC algorithms, some of them despite the simplicity of linear model, achieve performance comparable with deep SC algorithms. In the case of the MNIST dataset, the NSN SC algorithm [67] achieves in AA subband accuracy of 77.07% and NMI of 73.28%, which is better than the reported performance of several deep SC algorithms. In the case of the USPS dataset, the S0L0 LRSSC algorithm [10] achieves in the A subband accuracy of 83.06% and NMI of 80.38%, which is better or comparable to deep SC algorithms. In the case of the EYaleB dataset, the SSC algorithm [9], after IPD correction, in the D sub-band achieves an accuracy of 96.92% and NMI of 98.33%. That is basically the second best result together with the TSC SSC and TSC LLMC algorithms [12] and the DASC algorithm [21]. In the case of the ORL dataset, the GMC LRSSC algorithm [10] in the original domain, after IPD correction, achieved an accuracy of 81.23% and NMI of 90.08%. That is better than performance of several deep SC algorithms. In the case of the COIL20 dataset, the NSN algorithm achieves in the AH sub-band, after IPD correction, an accuracy of 81.56% and NMI of 86.88%. That is better than TSC LRR [12] and deep SC algorithm LBDR [25]. In the case of the COIL100 dataset, the RTSC and NSN algorithms achieve in AA sub-band accuracy of 68.78% and 79.64%,, and NMI of 86.81% and 91.32%, in respective order. In the case of the RTSC that is comparable to the performance of most deep SC algorithms cited in Table IX. In the case of the NSN that is better than the performance of deep SC algorithms cited in Table IX.

We briefly present the geometric interpretation of the clustering results described above, which are announced in Section III.D. For that purpose, we stay focused on affinities in the ambient and WP domain estimated from partitions obtained by the SSC algorithm. To make interpretation more convincing, we convert average affinities (22) and (23) into average angles between subspaces, i.e. $\hat{\phi}(\mathbf{X}) = \cos^{-1}\left(affinity(\mathbf{X})\right)$, and $\hat{\phi}\left(\mathbf{X}^{k^*}\right) = \cos^{-1}\left(affinity\left(\mathbf{X}^{k^*}\right)\right)$. For MNIST, USPS, and COIL20 datasets corresponding values of $\left(\hat{\phi}(\mathbf{X}), \hat{\phi}\left(\mathbf{X}^{k^*}\right)\right)$ are respectively given as: $(62.22^\circ, 49.00^\circ)$, $(50.52^\circ, 47.73^\circ)$, and $(67.43^\circ, 64.84^\circ)$. Since the best subbands were AA, A and A, the subspaces are closer in the WP domain. For EYaleB, ORL and COIL100 datasets corresponding angles are respectively given as: $(53.91^\circ, 78.48^\circ)$, $(58.52^\circ, 60.64^\circ)$ and $(68.43^\circ, 72.69^\circ)$. Since the best subbands were D, AH and AH, the subspaces moved far away from each other in the WP domain.

Another interesting result is the optimal value of subspace dimension *d* used in IPD correction. We remind that the expected dimensions according to the literature for MNIST and USPS datasets are 12, and for ORL, EYaleB, COIL20 and COIL100 datasets, are 9. However, in the case of the MNIST dataset, SSC [9] and RTSC [11] achieve the best performance in the AA subband after IPD correction with *d*=6. In the case of the ORL dataset, the WP MERA achieves the best performance after IPD correction with *d*=7, and SSC [9] achieves the best performance in the AH subband after IPD correction with *d*=4. In the case of the COIL20 dataset, RTSC [11] in the AH subband achieves the best performance after IPD correction with *d*=6. Finally, in the case of the COIL100 dataset, the SSC algorithm [9] in the AH subband achieves best performance after IPD correction with *d*=3. Comparison of (2) and (16) suggest that the noise suppression capability of the AA and the AH sub-bands reduces subspace dimension in the WP domain.

TABLE IV

CLUSTERING RESULTS ON MNIST DATASET. ALL WPD-SC RESULTS WERE OBTAINED IN AA SUBBAND.

| Algorithm | ACC [%] in-sample out-of-sample | NMI [%] in-sample out-of-sample | Rand [%] in-sample out-of-sample | F_score[%] in-sample out-of-sample | Purity [%] in-sample out-of-sample | Average affinity |
|---|---|---|---|---|---|---|
| WP MERA [27] | **99.33±2.38** **88.77±2.26** | **99.32±1.08** **80.20±1.79** | **99.03±2.38** **77.04±2.72** | **99.12±2.13** **79.31±2.42** | **99.45±1.71** **88.85±1.84** | |
| SSC [9] | 60.25±4.89 60.30±4.35 | 62.10±2.95 60.40±2.36 | 45.55±4.43 46.78±3.95 | 51.31±3.86 52.52±3.43 | 64.49±3.52 64.83±3.08 | 0.4814±0.0074 |
| IPD [35], d=10 | 60.89±4.31 61.13±4.06 | 62.59±2.98 60.54±2.32 | 46.04±4.14 47.14±3.75 | 51.73±3.64 52.83±3.26 | 64.94±3.20 65.20±2.91 | 0.4834±0.0078 |
| WP SSC | 64.28±3.97 63.36±3.46 | 65.62±2.97 62.38±2.38 | 50.73±4.07 50.61±3.42 | 55.84±3.60 55.78±3.02 | 68.50±3.06 68.04±2.63 | 0.6549±0.0069 |
| IPD, d=6 | 64.37±3.45 63.32±2.94 | 66.52±2.49 62.71±1.81 | 51.46±3.30 50.63±2.17 | 56.51±2.93 56.03±2.17 | 68.81±2.75 68.17±2.14 | 0.6544±0.0081 |
| S0L0 LRSSC [10] | 64.25±4.37 63.82±3.68 | 64.12±2.97 64.27±2.18 | 49.38±4.19 49.94±3.30 | 54.58±3.73 55.20±2.91 | 67.94±3.42 68.02±2.82 | 0.4840±0.0081 |
| WP S0L0 LRSSC | 69.69±4.32 68.65±3.96 | 68.32±2.62 64.44±2.03 | 55.75±3.89 55.36±3.31 | 60.25±3.47 50.98±2.95 | 72.16±3.10 71.18±2.62 | 0.6623±0.0060 |
| IPD, d=10 | 70.38±4.16 69.17±3.83 | 68.58±2.68 64.55±2.20 | 56.30±4.18 55.66±3.49 | 60.73±3.72 60.22±3.10 | 72.56±2.99 71.34±2.52 | 0.6630±0.0056 |
| LRR [8] | 54.46±5.40 54.60±5.16 | 60.71±3.27 58.89±2.80 | 38.51±5.20 40.89±4.79 | 45.78±4.35 47.75±4.03 | 60.30±4.17 61.06±3.90 | 0.4493±0.0132 |
| IPD, d=11 | 60.69±5.38 60.53±4.94 | 64.18±3.56 61.80±2.89 | 46.39±5.37 47.21±4.54 | 52.42±4.64 53.21±3.89 | 64.52±4.68 64.67±4.31 | 0.5180±0.0501 |
| WP LRR | 57.48±5.88 57.53±5.84 | 62.87±3.42 60.41±2.84 | 40.53±5.41 43.09±4.89 | 47.52±4.55 49.65±4.14 | 62.97±4.25 63.54±4.02 | 0.6397±0.0108 |
| IPD, d=11 | 65.73±5.82 65.52±5.26 | 68.76±3.84 64.54±3.23 | 53.16±5.91 52.63±5.04 | 58.38±5.09 57.94±4.33 | 68.86±4.94 68.30±4.55 | 0.6936±0.0403 |
| NSN [67] | 68.82±5.02 68.42±4.75 | 66.73±3.76 64.85±2.86 | 53.75±5.24 54.83±4.58 | 58.52±4.68 59.57±4.08 | 71.48±4.24 71.53±3.71 | 0.4848±0.0093 |
| WP NSN | 77.07±5.75 75.83±5.24 | 73.28±3.66 69.37±2.78 | 62.83±5.24 61.99±4.76 | 66.63±5.20 65.94±4.23 | 78.42±4.59 77.31±3.88 | 0.6561±0.0078 |
| RTSC [11] | 60.49±3.70 60.43±2.97 | 61.81±2.96 60.71±2.34 | 45.86±4.16 47.51±3.41 | 51.48±3.68 53.03±3.01 | 65.16±3.24 65.84±2.70 | 0.4741±0.0077 |
| IPD, d=8 | 63.06±4.95 62.42±4.37 | 66.53±3.22 63.70±2.38 | 50.60±4.99 50.67±4.10 | 55.86±4.36 55.99±3.55 | 68.09±3.97 68.02±3.29 | 0.4643±0.0100 |
| WP RTSC | 64.66±4.21 63.69±3.70 | 69.69±3.10 65.77±2.14 | 53.79±4.55 53.05±3.64 | 58.79±3.97 58.16±3.16 | 69.85±3.41 69.44±2.91 | 0.6401±0.0082 |
| IPD, d=6 | 66.11±4.36 64.80±3.86 | 70.90±2.82 66.67±1.97 | 55.56±4.40 54.40±3.57 | 60.33±3.84 59.34±3.08 | 71.38±3.51 70.60±3.00 | 0.6417±0.0078 |
| | | **Deep       networks** | | | | |
| DSLSP [69] | 91.64 | 82.96 | - | - | - | - |
| AASSC [19] | 84.60 | 76.09 | | | 84.60 | |
| DEC [71] | 61.20 | 57.53 | - | - | 63.20 | - |
| DSC-L2 [17], reported from [19] | 75.00 | 73.19 | - | - | 79.91 | - |
| DASC [21] | 80.40 | 78.00 | - | - | 83.70 | - |
| PSSC [22] | 84.30 | 76.76 | - | - | 84.30 | - |
| MESC [18] | 81.11 | 82.26 | - | - | - | - |

TABLE V

CLUSTERING RESULTS ON USPS DATASET. ALL WPD-SC RESULTS WERE OBTAINED IN A SUBBAND.

| Algorithm | ACC [%] in-sample out-of-sample | NMI [%] in-sample out-of-sample | Rand [%] in-sample out-of-sample | F_score[%] in-sample out-of-sample | Purity [%] in-sample out-of-sample | Average affinity |
|---|---|---|---|---|---|---|
| WP MERA [27] | **99.70±0.26** **92.12±1.19** | **99.39±0.52** **86.00±1.84** | **99.34±0.57** **83.45±2.34** | **99.41±0.51** **85.08±2.11** | **99.70±0.26** **92.12±1.19** | |
| SSC [9] | 75.11±5.72 78.19±5.03 | 73.74±3.07 74.02±2.39 | 62.69±5.20 70.19±4.56 | 66.64±4.53 73.62±3.97 | 76.67±4.86 79.75±4.05 | 0.6358±0.0073 |
| IPD [35], d=11 | 75.45±5.68 78.69±4.57 | 74.95±3.24 75.26±2.39 | 64.19±5.51 71.75±4.17 | 68.01±4.81 75.02±3.63 | 77.21±4.85 80.34±3.87 | 0.6356±0.0065 |
| WP SSC | 75.39±5.78 78.23±5.09 | 73.75±2.85 73.86±2.48 | 62.84±4.93 70.19±4.64 | 66.64±4.30 73.61±4.06 | 76.98±4.77 79.87±3.89 | 0.6726±0.0074 |
| S0L0 LRSSC [10] | 82.75±6.14 83.64±5.60 | 80.57±3.09 78.95±2.34 | 72.69±5.65 76.56±5.29 | 75.47±5.03 79.13±4.73 | 84.37±4.53 85.93±3.14 | 0.6376±0.0056 |
| WP S0L0 LRSSC | 83.06±5.40 83.68±5.43 | 80.38±2.67 78.88±2.15 | 72.69±4.82 76.66±4.88 | 75.47±4.29 79.23±4.35 | 84.58±3.90 85.85±2.99 | 0.6752±0.0051 |
| LRR [8] | 69.24±3.66 72.25±3.47 | 68.90±2.78 70.84±2.13 | 50.20±5.30 58.75±4.19 | 55.84±4.54 63.51±3.58 | 72.59±3.04 77.77±2.21 | 0.6411±0.0060 |
| IPD, d=12 | 80.77±5.67 82.02±5.50 | 80.76±2.41 79.18±2.10 | 71.09±4.21 75.74±5.08 | 74.18±4.34 78.49±4.51 | 82.08±4.59 83.97±3.65 | 0.6628±0.0344 |
| WP LRR | 69.58±4.17 72.50±3.90 | 69.40±2.65 71.20±2.10 | 50.99±4.87 59.42±3.82 | 56.43±4.20 64.08±3.31 | 72.97±3.26 77.92±2.37 | 0.6763±0.0060 |
| IPD, d=12 | 79.87±5.36 81.68±5.01 | 80.57±2.42 78.92±2.08 | 70.29±4.85 75.66±4.49 | 73.50±4.23 78.46±3.96 | 80.69±4.71 82.85±3.67 | 0.7108±0.0360 |
| NSN [67] | 74.67±5.40 77.13±5.84 | 70.24±3.50 72.31±2.81 | 58.32±5.20 66.14±5.57 | 63.04±4.83 70.05±5.25 | 76.75±4.07 80.50±3.25 | 0.6511±0.0055 |
| WP NSN | 74.39±5.32 77.08±5.62 | 69.83±3.37 71.95±2.72 | 58.59±5.79 66.45±4.94 | 62.66±4.95 69.96±4.94 | 76.30±4.04 80.14±3.16 | 0.6870±0.0055 |
| RTSC [11] | 72.11±5.97 75.37±5.46 | 69.54±3.63 71.50±2.73 | 58.08±5.41 65.20±5.39 | 62.34±4.82 68.97±4.83 | 75.31±4.20 79.09±3.26 | 0.6557±0.0056 |
| IPD, d=12 | 72.35±5.72 75.10±5.32 | 70.31±3.34 71.84±2.42 | 58.78±5.24 65.27±4.90 | 62.97±4.68 69.03±4.39 | 75.57±4.23 79.12±3.21 | 0.6437±0.0053 |
| WP RTSC | 71.65±5.42 75.36±5.13 | 69.72±3.47 71.63±2.71 | 58.19±5.17 65.69±5.16 | 62.45±4.61 69.41±4.63 | 75.14±3.97 79.10±2.99 | 0.6807±0.0055 |
| IPD, d=12 | 71.42±5.00 74.91±4.82 | 70.02±3.13 71.64±2.43 | 58.34±4.75 65.49±4.85 | 62.58±4.24 69.24±4.36 | 75.09±3.72 78.90±2.83 | 0.6804±0.0052 |
| **Deep networks** | | | | | | |
| DSLSP [69] | 83.29 | 83.70 | - | - | - | - |
| DSC-L1 [17], reported from [19] | 79.65 | 82.95 | - | - | - | - |
| MESC [18] | 81.49 | 86.34 | - | - | - | - |

TABLE VI

CLUSTERING RESULTS ON EYALEB DATASET. SUBBAND FOR EACH WPD-SC ALGORITHM IS REPORTED IN TABLE.

| Algorithm | ACC [%] in-sample out-of-sample | NMI [%] in-sample out-of-sample | Rand [%] in-sample out-of-sample | F_score[%] in-sample out-of-sample | Purity [%] in-sample out-of-sample | Average affinity |
|---|---|---|---|---|---|---|
| WP MERA [27] | **99.49±1.31** **92.93±1.39** | **99.89±0.28** **92.83±1.03** | **99.51±1.25** **86.17±2.07** | **99.52±1.21** **86.52±2.02** | **99.60±1.02** **93.00±1.26** | |
| SSC [9] | 75.65±1.84 81.36±2.17 | 80.43±1.65 85.79±1.67 | 38.59±4.66 57.45±3.97 | 40.77±4.39 58.73±3.81 | 76.29±1.67 82.01±2.05 | 0.5890±0.0038 |
| IPD [35], d=3 | 87.64±2.14 86.54±2.18 | 91.32±0.78 90.49±1.05 | 80.50±2.11 77.46±2.91 | 81.02±2.05 78.06±2.82 | 88.12±1.92 86.96±2.04 | 0.6006±0.0065 |

| Algorithm | | | | | | |
|---|---|---|---|---|---|---|
| WP-D SSC | 95.01±2.58 | 98.05±0.72 | 94.10±2.70 | 94.26±2.63 | 95.92±2.01 | 0.1996±0.0022 |
| | 91.49±2.55 | 94.05±0.92 | 86.71±2.76 | 87.06±2.68 | 92.26±2.06 | |
| IPD, d=8 | 96.92±1.68 | 98.63±0.42 | 96.24±1.60 | 96.34±1.55 | 97.46±1.27 | 0.2007±0.0019 |
| | 93.07±1.62 | 94.45±0.71 | 88.38±1.71 | 88.67±1.66 | 93.57±1.24 | |
| GMC LRSSC [10] | 88.69±1.73 | 91.10±1.11 | 78.36±3.00 | 78.95±2.91 | 89.96±1.63 | 0.5943±0.0073 |
| | 87.06±1.67 | 89.83±1.13 | 78.49±2.77 | 79.05±2.69 | 87.37±1.54 | |
| IPD, d=8 | 89.27±1.21 | 91.82±0.68 | 81.21±1.95 | 81.72±1.89 | 89.62±1.53 | 0.6080±0.0050 |
| | 88.65±1.90 | 91.47±0.89 | 80.02±2.30 | 80.54±2.23 | 89.01±1.74 | |
| WP-DH GMC LRSSC | 87.74±1.62 | 90.85±0.73 | 76.98±2.60 | 77.62±2.52 | 88.31±1.32 | 0.2125±0.0013 |
| | 87.95±1.90 | 91.30±0.99 | 77.48±3.01 | 78.09±2.92 | 88.55±1.63 | |
| IPD, d=9 | 88.25±1.97 | 90.96±0.78 | 77.79±2.57 | 78.40±2.49 | 89.16±1.28 | 0.2129±0.0009 |
| | 88.49±1.82 | 91.50±0.88 | 78.27±2.83 | 78.85±2.75 | 89.90±1.67 | |
| NSN [67] | 72.98±2.53 | 75.35±1.46 | 56.65±2.58 | 57.72±2.51 | 73.48±2.39 | 0.5689±0.0072 |
| | 74.86±2.12 | 79.81±1.16 | 60.07±2.23 | 61.12±2.17 | 75.40±2.04 | |
| WP-DH NSN | 88.08±1.66 | 89.75±0.75 | 80.19±1.80 | 80.71±1.75 | 88.28±1.57 | 0.2122±0.0011 |
| | 87.32±1.73 | 89.78±0.94 | 77.76±2.30 | 78.34±2.23 | 87.61±1.65 | |
| RTSC [11] | 40.50±1.62 | 52.26±1.17 | 17.83±1.51 | 20.53±1.38 | 42.56±1.36 | 0.4889±0.0055 |
| | 46.25±2.06 | 59.92±1.29 | 24.73±2.09 | 27.00±1.97 | 47.96±1.88 | |
| WP-DH RTSC | 87.71±1.38 | 89.81±0.77 | 74.29±2.46 | 75.00±2.38 | 87.97±1.15 | 0.2118±0.0009 |
| | 88.14±1.54 | 90.75±0.89 | 76.26±2.78 | 76.89±2.70 | 88.54±1.34 | |
| $S_{2/3}$-LRR [68] | 68.54±1.98 | 74.90±1.19 | 52.50±1.86 | 53.84±1.79 | 69.03±1.89 | 0.5659±0.0091 |
| | 70.94±2.35 | 79.43±1.29 | 57.43±2.82 | 58.59±2.72 | 71.55±2.15 | |
| IPD, d=9 | 91.56±1.55 | 92.93±0.61 | 83.82±1.77 | 84.25±1.71 | 91.71±1.43 | 0.6124±0.0033 |
| | 90.81±1.58 | 92.56±0.85 | 82.56±2.37 | 83.01±2.31 | 91.01±1.48 | |
| WP-D $S_{2/3}$-LRR | 69.59±1.94 | 75.10±1.31 | 48.65±2.45 | 50.16±2.34 | 70.28±1.34 | 0.2317±0.0020 |
| | 72.98±2.06 | 79.67±1.36 | 54.86±3.40 | 56.13±3.27 | 73.74±1.87 | |
| IPD, d=7 | 88.12±1.70 | 90.89±0.76 | 77.78±2.48 | 78.39±2.40 | 88.51±1.41 | 0.1939±0.0012 |
| | 87.87±1.95 | 91.09±1.01 | 77.42±2.85 | 78.02±2.77 | 88.37±1.64 | |
| **Deep networks** | | | | | | |
| TSC SSC [12] | 99 | 94 | 96 | 98 | 98 | |
| TSC LRR [12] | 69 | 74 | 75 | 74 | 72 | |
| DSLSP [69] | 97.62 | 96.74 | - | - | - | - |
| DSC-L2 [17], reported from [19] | 97.73 | 97.03 | - | - | - | - |
| DASC [21] | 98.56 | 98.01 | - | - | - | - |
| MESC [18] | 98.03 | 97.27 | - | - | - | - |
| SAE [33] | 88.75 | 87.53 | - | - | - | - |
| DCSC [24] | 92.36 | 94.27 | - | - | - | - |
| LBDR [25] | 84.73 | 86.75 | - | - | - | - |

TABLE VII

CLUSTERING RESULTS ON ORL DATASET. SUBBAND FOR EACH WPD-SC ALGORITHM IS REPORTED IN TABLE.

| Algorithm | ACC [%] in-sample out-of-sample | NMI [%] in-sample out-of-sample | Rand [%] in-sample out-of-sample | F_score[%] in-sample out-of-sample | Purity [%] in-sample out-of-sample | Average affinity |
|---|---|---|---|---|---|---|
| WP MERA [27] | 81.71±2.83 | 91.28±1.36 | 73.47±3.64 | 74.08±3.56 | 84.08±2.47 | |
| | 80.31±3.36 | 92.31±1.34 | 66.02±5.02 | 66.68±4.91 | 81.98±3.06 | |
| IPD [35], d=5 | 88.98±2.69 | 94.02±1.27 | 81.95±3.67 | 82.35±3.59 | 89.98±2.30 | |
| | 86.72±2.98 | 94.41±1.26 | 74.95±4.94 | 75.41±4.84 | 87.56±2.80 | |
| SSC [9] | 72.23±2.93 | 86.35±1.32 | 59.66±3.57 | 60.61±3.48 | 75.27±2.53 | 0.5220±0.0137 |
| | 70.06±3.27 | 87.46±1.57 | 48.74±5.46 | 49.82±5.31 | 72.13±2.96 | |

| | | | | | |
|---|---|---|---|---|---|
| IPD, d=7 | 75.31±3.09 | 88.04±1.38 | 64.07±3.77 | 64.91±3.67 | 78.45±2.44 | 0.5090±0.0117 |
| | 72.16±3.13 | 88.39±1.40 | 51.80±4.71 | 52.80±4.58 | 74.38±2.87 | |
| WP-AH SSC | 73.72±2.66 | 86.87±1.23 | 61.67±3.20 | 62.56±3.12 | 76.64±2.24 | 0.4993±0.0051 |
| | 69.71±3.12 | 87.02±1.39 | 47.84±4.55 | 48.92±4.42 | 71.93±2.92 | |
| IPD, d=4 | 76.14±3.07 | 88.63±1.50 | 65.78±4.05 | 65.58±3.95 | 79.47±2.73 | 0.4146±0.0055 |
| | 72.21±3.00 | 88.18±1.50 | 51.41±5.09 | 52.41±4.96 | 74.57±2.85 | |
| GMC LRSSC [10] | 77.97±2.10 | 88.45±1.29 | 67.12±3.35 | 67.81±3.27 | 79.84±1.21 | 0.3281±0.0052 |
| | 75.14±2.87 | 89.28±1.39 | 55.86±5.14 | 56.73±5.01 | 76.74±2.72 | |
| IPD, d=5 | 81.23±2.64 | 90.08±1.17 | 71.25±3.16 | 71.90±3.08 | 82.90±2.15 | 0.5204±0.0100 |
| | 77.22±3.05 | 90.29±1.27 | 59.37±4.58 | 60.16±4.47 | 78.74±2.77 | |
| WP-A GMC LRSSC | 78.61±2.38 | 89.00±1.18 | 68.09±3.09 | 68.81±3.02 | 80.59±2.11 | 0.5370±0.0098 |
| | 76.07±2.81 | 89.94±1.20 | 58.02±4.30 | 58.85±4.20 | 77.73±2.54 | |
| IPD, d=4 | 81.01±2.47 | 90.30±1.18 | 71.02±3.16 | 72.02±3.16 | 82.89±2.13 | 0.5822±0.0092 |
| | 77.64±2.70 | 90.71±1.36 | 60.99±4.84 | 61.75±4.73 | 79.26±2.62 | |
| NSN [67] | 67.80±2.52 | 82.78±1.27 | 51.98±3.21 | 53.11±3.12 | 70.49±2.13 | 0.3187±0.0055 |
| | 65.05±2.86 | 84.65±1.29 | 41.18±3.98 | 42.43±3.87 | 67.33±2.79 | |
| WP-AH NSN | 69.26±2.64 | 83.77±1.33 | 55.27±3.18 | 56.34±3.09 | 72.05±2.30 | 0.4048±0.0080 |
| | 67.92±3.36 | 85.88±1.59 | 45.59±4.94 | 46.73±4.81 | 69.76±3.25 | |
| RTSC [11] | 69.12±2.70 | 82.86±1.40 | 53.31±3.43 | 54.38±3.34 | 71.65±2.33 | 0.4870±0.0065 |
| | 66.57±2.89 | 85.27±1.40 | 43.30±4.35 | 44.48±4.23 | 68.77±2.88 | |
| WP-AH RTSC | 69.99±2.75 | 82.84±1.42 | 54.00±3.52 | 55.05±3.43 | 72.45±2.50 | 0.3798±0.0039 |
| | 69.47±2.33 | 86.35±1.33 | 46.95±4.11 | 48.02±4.01 | 71.50±2.25 | |
| $S_{2/3}$-LRR [68] | 68.00±3.00 | 82.88±1.47 | 53.57±3.53 | 54.63±3.45 | 70.99±2.63 | 0.5093±0.0117 |
| | 67.33±3.31 | 85.90±1.43 | 45.42±4.63 | 46.53±4.51 | 69.72±3.02 | |
| IPD d=6 | 75.76±2.62 | 86.92±1.26 | 63.44±3.21 | 64.27±3.13 | 77.91±2.17 | 0.5054±0.0122 |
| | 73.57±2.86 | 88.61±1.21 | 53.69±4.43 | 54.62±4.32 | 75.33±2.69 | |
| WP-AH $S_{2/3}$-LRR | 68.86±2.82 | 83.18±1.49 | 54.54±3.44 | 55.58±3.36 | 71.86±2.55 | 0.4053±0.0062 |
| | 68.43±3.23 | 86.22±1.64 | 46.51±5.16 | 47.59±5.04 | 70.05±3.04 | |
| IPD d=5 | 81.55±2.59 | 90.07±1.23 | 71.47±3.24 | 72.11±3.16 | 83.25±2.25 | 0.3956±0.0063 |
| | 77.78±3.02 | 90.08±1.30 | 59.43±4.66 | 60.21±4.56 | 78.99±2.78 | |
| **Deep networks** | | | | | | |
| DSLSP [69] | 87.55 | 92.49 | - | - | - | - |
| AASSC [19] | **90.75** | **94.31** | - | - | 91.75 | |
| DSC-L2 [17], reported from [19] | 86.00 | 90.34 | - | - | - | - |
| DASC [21] | 88.25 | 93.15 | - | - | 89.25 | - |
| PSSC [22] | 86.75 | 93.49 | - | - | 89.25 | - |
| MESC [18] | **90.25** | **93.59** | - | - | - | - |
| SAE [23] | 74.81 | 88.0 | - | - | - | |
| DCSC [24] | 83.52 | 90.1 | - | - | - | - |
| LBDR [25] | 77.68 | 89.12 | - | - | - | - |

TABLE VIII

CLUSTERING RESULTS ON COIL20 DATASET. SUBBAND FOR EACH WPD-SC ALGORITHM IS REPORTED IN TABLE.

| Algorithm | ACC [%] in-sample out-of-sample | NMI [%] in-sample out-of-sample | Rand [%] in-sample out-of-sample | F_score[%] in-sample out-of-sample | Purity [%] in-sample out-of-sample | Average affinity |
|---|---|---|---|---|---|---|
| WP MERA [27] | 96.04±4.59 | 98.82±1.33 | 94.95±4.56 | 96.24±4.33 | 97.02±3.41 | |
| | 94.19±4.45 | 96.48±1.53 | 92.34±4.42 | 92.73±4.18 | 95.04±3.48 | |
| IPD [35], d=10 | **99.94±0.15** | **99.93±0.18** | **99.88±0.31** | **99.88±0.29** | **99.94±0.15** | |
| | **98.02±0.72** | **97.63±0.77** | **96.02±1.39** | **96.22±1.32** | **98.02±0.72** | |
| SSC [9] | 70.55±3.40 | 82.44±1.69 | 61.63±3.51 | 63.66±3.29 | 74.14±2.60 | 0.3838±0.0102 |
| | 68.99±3.40 | 81.10±1.61 | 69.37±3.35 | 62.54±3.12 | 72.28±3.60 | |
| IPD, d=8 | 75.93±2.88 | 85.71±1.45 | 68.49±3.44 | 70.12±3.23 | 78.98±2.16 | 0.3909±0.0063 |
| | 74.04±2.79 | 84.29±1.54 | 66.91±3.63 | 68.68±3.39 | 76.85±2.19 | |
| WP-A SSC | 72.03±3.31<br>70.00±3.32 | 82.95±1.79<br>81.53±1.81 | 63.27±3.94<br>61.77±3.96 | 65.18±3.69<br>63.83±3.69 | 75.10±2.77<br>73.05±2.80 | 0.4252±0.0092 |
| IPD, d=8 | 76.53±3.11<br>74.71±2.92 | 86.13±1.65<br>84.75±1.87 | 69.43±3.50<br>67.96±3.53 | 70.99±3.29<br>69.66±3.32 | 79.84±2.36<br>77.75±2.46 | 0.4293±0.0063 |
| GMC LRSSC [10] | 71.45±2.70<br>69.93±2.39 | 82.98±2.41<br>81.70±1.44 | 61.37±3.61<br>60.04±3.64 | 63.46±3.35<br>62.28±3.36 | 75.02±2.12<br>73.06±2.00 | 0.3733±0.0073 |
| IPD, d=6 | 71.71±2.58<br>70.23±2.51 | 83.01±1.52<br>81.70±1.64 | 61.64±3.51<br>60.85±3.42 | 63.71±3.27<br>63.02±3.18 | 75.11±2.06<br>73.33±2.06 | 0.4055±0.0076 |
| WP-AH GMC LRSSC | 75.59±2.44<br>74.44±2.10 | 83.30±1.37<br>82.02±1.33 | 67.12±2.72<br>66.41±2.52 | 68.76±2.57<br>68.14±2.37 | 77.83±1.99<br>76.44±1.72 | 0.3159±0.0039 |
| IPD, d=6 | 76.19±2.33<br>74.86±1.97 | 83.80±1.36<br>82.36±1.32 | 67.63±2.60<br>66.85±2.44 | 69.26±2.45<br>68.54±2.29 | 78.40±1.86<br>77.03±1.57 | 0.2928±0.0049 |
| LRR [8] | 61.30±4.42<br>59.32±4.34 | 75.57±2.01<br>74.48±2.04 | 49.77±2.09<br>49.12±2.03 | 52.58±4.66<br>52.07±4.59 | 64.77±3.55<br>62.33±3.65 | 0.3771±0.0079 |
| IPD, d=7 | 70.85±3.87<br>69.34±3.62 | 82.97±1.73<br>81.46±1.70 | 60.59±4.86<br>59.39±4.45 | 62.73±4.50<br>61.67±4.11 | 74.66±2.88<br>72.65±2.70 | 0.3790±0.0079 |
| WP-AH LRR | 70.76±3.00<br>69.82±3.04 | 80.68±1.44<br>79.86±1.43 | 61.55±3.06<br>61.90±3.00 | 63.55±2.86<br>63.93±2.80 | 73.90±3.49<br>72.80±2.62 | 0.3173±0.0048 |
| IPD, d=10 | 72.90±3.10<br>71.65±2.87 | 81.63±1.55<br>80.36±1.73 | 62.14±3.29<br>61.60±3.03 | 64.14±3.93<br>63.67±2.84 | 75.79±2.23<br>74.26±2.19 | 0.3270±0.0077 |
| NSN [67] | 74.02±3.24<br>72.17±3.28 | 83.50±1.60<br>81.74±1.74 | 65.93±3.27<br>64.57±3.20 | 67.69±3.08<br>66.44±3.01 | 76.30±2.85<br>74.32±2.82 | 0.4228±0.0128 |
| WP-AH NSN | 75.53±3.80<br>74.32±3.64 | 85.09±1.83<br>83.09±1.85 | 68.64±3.89<br>66.92±3.58 | 70.25±3.67<br>68.66±3.37 | 78.00±3.13<br>76.01±3.02 | 0.3026±0.0065 |
| IPD, d=10 | 81.56±2.40<br>79.34±2.10 | 86.88±1.46<br>85.22±1.42 | 74.16±2.81<br>72.29±2.45 | 75.43±2.66<br>73.69±2.32 | 82.74±2.16<br>80.58±1.95 | 0.3253±0.0035 |
| RTSC [11] | 72.51±3.29<br>70.79±2.89 | 82.23±1.49<br>80.71±1.43 | 63.14±3.42<br>61.89±3.07 | 65.04±3.21<br>63.93±2.87 | 75.55±2.43<br>73.55±2.18 | 0.3778±0.0062 |
| IPD, d=5 | 73.94±3.23<br>72.16±2.87 | 82.92±1.54<br>81.13±1.45 | 65.07±3.36<br>63.81±3.01 | 66.86±3.16<br>65.71±2.82 | 76.83±2.48<br>74.75±2.12 | 0.4295±0.0074 |
| WP-AH RTSC | 74.84±3.15<br>73.64±2.91 | 83.89±1.42<br>82.51±1.44 | 65.36±3.83<br>64.56±3.56 | 67.16±3.58<br>66.45±3.32 | 78.20±2.15<br>76.72±2.01 | 0.3123±0.0047 |
| IPD, d=6 | 76.25±3.21<br>75.17±2.92 | 84.82±1.47<br>83.44±1.52 | 67.37±3.66<br>66.84±3.26 | 69.05±3.43<br>68.59±3.05 | 79.17±2.31<br>77.84±2.22 | 0.2910±0.0039 |
| | **Deep** | **networks** | | | | |
| TSC LLMC [12] | 98 | 94 | 94 | 97 | 99 | |
| TSC LRR [12] | 78 | 83 | 72 | 74 | 72 | |
| DSLSP [69] | 97.57 | 97.40 | - | - | - | - |
| AASSC [19] | 98.40 | 98.29 | | | 98.40 | |
| DSC-L2 [17], reported from [19] | 93.68 | 94.08 | - | - | 93.97 | - |
| DASC [21] | 96.39 | 96.86 | - | - | 96.32 | - |
| MESC [18] | 98.40 | 98.29 | - | - | 98.40 | - |
| SAE [23] | 86.29 | 90.28 | - | - | - | - |
| DCSC [24] | 92.08 | 95.39 | - | - | - | - |
| LBDR [25] | 78.59 | 86.97 | - | - | - | - |

TABLE IX

CLUSTERING RESULTS ON COIL100 DATASET. SUBBAND FOR EACH WPD-SC ALGORITHM IS REPORTED IN TABLE.

| Algorithm | ACC [%] in-sample out-of-sample | NMI [%] in-sample out-of-sample | Rand [%] in-sample out-of-sample | F_score[%] in-sample out-of-sample | Purity [%] in-sample out-of-sample | Average affinity |
|---|---|---|---|---|---|---|
| WP MERA [27] | 84.59±1.85 80.01±1.80 | 94.25±0.51 90.73±0.61 | 80.40±1.83 72.80±1.89 | 80.60±1.81 73.07±1.87 | 88.61±1.47 81.86±1.52 | |
| IPD, d=10 | **87.45±1.49** **82.39±1.62** | **94.73±0.59** **91.18±0.64** | **82.82±1.98** **74.63±1.84** | **82.99±1.96** **74.88±1.82** | **88.89±1.48** **83.65±1.41** | |
| SSC [9] | 51.17±1.33 51.26±1.25 | 78.08±0.69 77.80±0.69 | 41.16±1.78 41.12±1.77 | 41.84±1.75 41.80±1.74 | 58.50±0.97 57.93±0.98 | 0.3676±0.0025 |
| IPD, d=2 | 64.57±1.87 61.64±1.68 | 85.83±0.66 82.14±0.66 | 55.39±2.35 51.91±2.06 | 55.89±2.31 52.45±2.03 | 69.07±1.59 66.14±1.37 | 0.6250±0.0030 |
| WP-AH SSC | 62.36±1.51 61.50±1.46 | 84.52±0.80 83.56±0.77 | 49.13±4.13 48.54±3.88 | 49.75±4.05 49.17±3.81 | 67.80±1.13 66.81±1.14 | 0.2975±0.0028 |
| IPD, d=3 | 67.24±1.57 65.20±1.50 | 87.64±0.63 84.68±0.67 | 55.26±3.67 53.64±2.47 | 55.80±3.60 54.19±2.42 | 71.82±1.30 69.64±1.21 | 0.2544±0.0030 |
| S0L0 LRSSC [10] | 50.47±1.19 49.86±1.18 | 75.52±0.40 75.44±0.40 | 43.51±1.01 43.13±1.06 | 44.09±1.00 43.72±1.04 | 53.64±0.90 52.87±0.91 | 0.3828±0.0022 |
| WP-AH S0L0 LRSSC | 54.96±1.39 54.32±1.32 | 79.53±0.59 79.09±0.56 | 46.62±1.65 47.01±1.47 | 47.23±1.62 47.60±1.45 | 60.51±1.01 59.58±1.03 | 0.3037±0.0025 |
| LRR [8] | 36.84±2.30 38.77±2.08 | 69.20±1.83 72.18±1.40 | 15.41±2.49 20.04±4.32 | 16.79±4.15 21.26±4.19 | 43.19±1.96 44.60±1.84 | 0.3525±0.0033 |
| IPD, d=9 | 56.73±1.47 57.29±1.46 | 82.67±0.48 83.13±0.52 | 45.95±2.47 47.24±1.99 | 46.62±2.42 47.86±1.96 | 61.63±1.06 61.98±1.22 | 0.3802±0.0069 |
| WP-AH LRR | 34.92±1.90 37.80±2.01 | 68.30±1.33 71.48±1.20 | 11.89±2.39 15.59±2.19 | 13.42±2.31 16.97±2.13 | 43.51±1.55 46.31±1.79 | 0.2751±0.0033 |
| IPD, d=4 | 58.24±2.14 58.53±2.03 | 81.30±0.83 81.59±0.73 | 33.21±4.94 40.71±2.48 | 34.22±4.82 41.48±2.42 | 64.10±1.66 64.69±1.56 | 0.2553±0.0039 |
| NSN [67] | 57.15±1.11 57.17±0.98 | 79.88±0.43 80.23±0.46 | 49.20±1.06 48.28±1.08 | 49.73±1.05 48.81±1.07 | 60.21±1.03 60.24±0.92 | 0.3718±0.0019 |
| WP-AA NSN | 79.64±1.42 78.25±1.21 | 91.32±0.40 90.63±0.46 | 74.28±1.34 71.93±1.30 | 74.54±1.33 72.21±1.28 | 81.12±1.21 79.74±1.10 | 0.3062±0.0022 |
| RTSC [11] | 55.83±1.68 54.88±1.65 | 84.37±0.41 84.54±0.50 | 49.98±1.73 49.33±1.70 | 50.58±1.70 49.91±1.67 | 66.36±1.12 66.42±1.23 | 0.3624±0.0044 |
| WP-AA RTSC | 68.78±1.07 68.61±1.27 | 86.81±0.27 86.77±0.39 | 62.04±1.10 60.85±1.35 | 62.45±1.08 61.26±1.33 | 72.90±0.91 72.55±0.87 | 0.5040±0.0028 |
| **Deep networks** | | | | | | |
| MAESC [18] | 71.88 | 90.76 | | | | |
| DSLSP [70], reported from [18] | 65.86 | 89.14 | - | - | - | - |
| DSC-L2 [17], reported from [18] | 67.71 | 89.08 | . | - | - | - |
| LRAE [70], reported from [18] | 56.62 | 79.77 | | | | |
| DSCNS S [71] | 71.42 | | | | | |
| DSRSCN [72] | 72.53 | 72.94 | | | | |
| DCFSC [73] | 72.70 | | | | | |

## V. CONCLUSIONS

Performance improvement of SC algorithms based on the union of subspaces model implies finding representation where subspaces are more separable. It is also essential to reduce the influence of data points near the intersection of subspaces. We proposed wavelet packets-based transformed domain subspace clustering to account for these issues. Depending on the number of resolution levels, wavelet packets yield several representations instantiated in terms of subbands. Since subbands are implemented through filtering, representations are complementary, and some suppress noise. Thus, a combination of original data with A, H, V, and D subbands yields complementary multi-view representation. The joint highly discriminative representation matrix used for clustering is learned by a low-rank MERA tensor network thanks to its capability to capture complex intra/inter-view dependencies in corresponding self-representation tensor. Wavelet packets are linear precomputed transforms implemented efficiently in terms of the filter bank, and the low-rank MERA approximation problem is itself based on a linear multi-view self-representation model. Despite that, clustering performance on six benchmark datasets achieved by the proposed approach outperformed, often by a large margin, performance achieved by deep SC algorithms. One possible explanation is that the complementarity of representations matters more than the linearity of the embedded space, which is the ultimate goal of deep SC algorithms. We also proposed to apply the existing linear single-view SC algorithm on the best subband selected during the validation phase. In most cases according to the Wilcoxon signed rank test, their clustering performance is improved significantly compared to the performance achieved on data in the original domain. Moreover, the performance of some algorithms such as NSN [67], S0L0/GMC LRSSC [10], and SSC [9] is comparable to the one achieved by some deep SC methods. Hence, these algorithms can be re-used with minimal cost in terms of pre-processing. Furthermore, due to practical reasons, procedure for clustering out-of-sample (test) data is proposed for wavelet packets MERA method as well as for linear single-view SC methods applied on the selected subband.

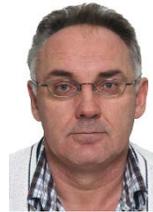

**Ivica Kopriva** (M'98-SM'04) received a Ph.D. degree in electrical engineering from the University of Zagreb, Croatia, in 1998. He was a senior research scientist with the ECE Department, The George Washington University, Washington, DC, USA, 2001-2005. Since 2006, he is with the Ruđer Bošković Institute, Zagreb, Croatia. His research is focused on unsupervised learning with applications in exploratory data analysis. He has been the recipient of the 2009 state award for Science of the Republic of Croatia.

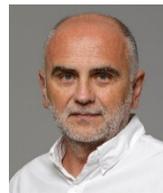

**Damir Seršić** (M'99) received the Diploma, Master, and Ph.D. degrees in technical sciences, electrical engineering, from the University of Zagreb, Zagreb, Croatia, in 1986, 1993, and 1999, respectively. He is promoted to full professor at the University of Zagreb Faculty of Electrical Engineering and Computing. Professor Seršić was a visiting researcher at the Colorado State University, Fort Collins, Colorado, USA, in 2012. His research interests include the theory and applications of wavelets, advanced signal and image processing, adaptive systems, blind source separation, and compressive sensing.

Supplement material for the paper "Subspace Clustering in Wavelet Packets Domain" by Ivica Kopriva and Damir Seršić

First row in each case stands for results obtained on in-sample data, while second row stands for results obtained on out-of-sample data.The best result is in bold font. The second best results is underlined. Wilcoxon test is used to verify whether subspace clustering (SC) in wavelet packets domain (WPD) on the best sub-band yielded statistically significantly improved performance. Cases with the *p*-values greater than 0.05 are shaded in grey. Except for the USPS dataset, SC in WPD yields significantly improved performance in the case of the majority of considered linear single-view SC algorithms. That is true for both in-sample and out-of-sample data.

TABLE IV: Clustering results on MNIST dataset. All WPD SC results were obtained in AA sub-band.

| Algorithm | ACC [%] in-sample out-of-sample | NMI [%] in-sample out-of-sample | Rand [%] in-sample out-of-sample | F_score[%] in-sample out-of-sample | Purity [%] in-sample out-of-sample | Average affinity |
|---|---|---|---|---|---|---|
| WP MERA [27] | **99.33±2.38** **88.77±2.26** | **99.32±1.08** **80.20±1.79** | **99.03±2.38** **77.04±2.72** | **99.12±2.13** **79.31±2.42** | **99.45±1.71** **88.85±1.84** | |
| SSC [9] | 60.25±4.89 60.30±4.35 | 62.10±2.95 60.40±2.36 | 45.55±4.43 46.78±3.95 | 51.31±3.86 52.52±3.43 | 64.49±3.52 64.83±3.08 | 0.4814±0.0074 |
| IPD [35], d=10 | 60.89±4.31 61.13±4.06 | 62.59±2.98 60.54±2.32 | 46.04±4.14 47.14±3.75 | 51.73±3.64 52.83±3.26 | 64.94±3.20 65.20±2.91 | 0.4834±0.0078 |
| WP SSC | 64.28±3.97 63.36±3.46 | 65.62±2.97 62.38±2.38 | 50.73±4.07 50.61±3.42 | 55.84±3.60 55.78±3.02 | 68.50±3.06 68.04±2.63 | 0.6549±0.0069 |
| IPD, d=6 | 64.37±3.45 63.32±2.94 | 66.52±2.49 62.71±1.81 | 51.46±3.30 50.86±2.47 | 56.51±2.93 56.03±2.17 | 68.81±2.75 68.17±2.14 | 0.6544±0.0081 |
| GMC LRSSC [10] | 62.39±4.22 62.10±3.67 | 62.96±2.87 61.05±2.17 | 47.33±3.73 48.29±3.22 | 52.76±3.30 53.77±2.84 | 66.32±2.98 66.53±2.63 | 0.4827±0.0081 |
| IPD, d=12 | 62.13±4.39 61.51±3.72 | 62.73±2.86 60.95±2.12 | 47.55±4.14 48.03±3.29 | 52.98±3.68 53.51±2.92 | 66.38±3.22 66.22±2.65 | 0.4801±0.0072 |
| WP GMC LRSSC | 67.69±3.86 66.99±3.14 | 64.39±2.85 61.52±1.80 | 51.76±3.83 51.97±2.76 | 56.62±3.43 56.91±2.46 | 69.48±2.98 68.85±2.11 | 0.6673±0.0058 |
| IPD, d=12 | 68.21±3.73 | 65.10±2.75 | 52.68±3.61 | 57.46±3.23 | 69.95±2.81 | 0.6656±0.0069 |

| | | | | | |
|---|---|---|---|---|---|
| | 67.22±3.20 | 62.02±1.91 | 52.69±2.84 | 57.56±2.54 | 69.19±2.07 | |
| S0L0 LRSSC [10] | 64.25±4.37 | 64.12±2.97 | 49.38±4.19 | 54.58±3.73 | 67.94±3.42 | 0.4840±0.0081 |
| | 63.82±3.68 | 64.27±2.18 | 49.94±3.30 | 55.20±2.91 | 68.02±2.82 | |
| WP S0L0 LRSSC | 69.69±4.32 | 68.32±2.62 | 55.75±3.89 | 60.25±3.47 | 72.16±3.10 | 0.6623±0.0060 |
| | 68.65±3.96 | 64.44±2.03 | 55.36±3.31 | 50.98±2.95 | 71.18±2.62 | |
| IPD, d=10 | 70.38±4.16 | 68.58±2.68 | 56.30±4.18 | 60.73±3.72 | 72.56±2.99 | 0.6630±0.0056 |
| | 69.17±3.83 | 64.55±2.20 | 55.66±3.49 | 60.22±3.10 | 71.34±2.52 | |
| LRR [8] | 54.46±5.40 | 60.71±3.27 | 38.51±5.20 | 45.78±4.35 | 60.30±4.17 | 0.4493±0.0132 |
| | 54.60±5.16 | 58.89±2.80 | 40.89±4.79 | 47.75±4.03 | 61.06±3.90 | |
| IPD, d=11 | 60.69±5.38 | 64.18±3.56 | 46.39±5.37 | 52.42±4.64 | 64.52±4.68 | 0.5180±0.0501 |
| | 60.53±4.94 | 61.80±2.89 | 47.21±4.54 | 53.21±3.89 | 64.67±4.31 | |
| WP LRR | 57.48±5.88 | 62.87±3.42 | 40.53±5.41 | 47.52±4.55 | 62.97±4.25 | 0.6397±0.0108 |
| | 57.53±5.84 | 60.41±2.84 | 43.09±4.89 | 49.65±4.14 | 63.54±4.02 | |
| IPD, d=11 | 65.73±5.82 | 68.76±3.84 | 53.16±5.91 | 58.38±5.09 | 68.86±4.94 | 0.6936±0.0403 |
| | 65.52±5.26 | 64.54±3.23 | 52.63±5.04 | 57.94±4.33 | 68.30±4.55 | |
| NSN [67] | 68.82±5.02 | 66.73±3.76 | 53.75±5.24 | 58.52±4.68 | 71.48±4.24 | 0.4848±0.0093 |
| | 68.42±4.75 | 64.85±2.86 | 54.83±4.58 | 59.57±4.08 | 71.53±3.71 | |
| WP NSN | 77.07±5.75 | 73.28±3.66 | 62.83±5.24 | 66.63±5.20 | 78.42±4.59 | 0.6561±0.0078 |
| | 75.83±5.24 | 69.37±2.78 | 61.99±4.76 | 65.94±4.23 | 77.31±3.88 | |
| RTSC [11] | 60.49±3.70 | 61.81±2.96 | 45.86±4.16 | 51.48±3.68 | 65.16±3.24 | 0.4741±0.0077 |
| | 60.43±2.97 | 60.71±2.34 | 47.51±3.41 | 53.03±3.01 | 65.84±2.70 | |
| IPD, d=8 | 63.06±4.95 | 66.53±3.22 | 50.60±4.99 | 55.86±4.36 | 68.09±3.97 | 0.4643±0.0100 |
| | 62.42±4.37 | 63.70±2.38 | 50.67±4.10 | 55.99±3.55 | 68.02±3.29 | |
| WP RTSC | 64.66±4.21 | 69.69±3.10 | 53.79±4.55 | 58.79±3.97 | 69.85±3.41 | 0.6401±0.0082 |
| | 63.69±3.70 | 65.77±2.14 | 53.05±3.64 | 58.16±3.16 | 69.44±2.91 | |
| IPD, d=6 | 66.11±4.36 | 70.90±2.82 | 55.56±4.40 | 60.33±3.84 | 71.38±3.51 | 0.6417±0.0078 |
| | 64.80±3.86 | 66.67±1.97 | 54.40±3.57 | 59.34±3.08 | 70.60±3.00 | |
| $S_{1/2}$-LRR [68] | 55.97±4.07 | 53.47±3.03 | 38.04±3.77 | 44.39±3.40 | 60.19±3.50 | 0.4957±0.0081 |
| | 57.35±3.54 | 55.60±2.76 | 41.95±3.58 | 48.01±3.21 | 62.22±3.28 | |
| IPD, d=9 | 58.63±4.16 | 59.58±3.07 | 43.38±4.24 | 49.38±3.73 | 63.49±3.44 | 0.4818±0.0096 |
| | 59.08±3.95 | 58.97±2.57 | 45.45±3.90 | 51.31±3.40 | 64.55±3.15 | |
| WP $S_{1/2}$-LRR | 59.09±3.84 | 55.27±3.27 | 41.63±4.07 | 47.52±3.65 | 62.92±3.52 | 0.6750±0.0056 |

| | 54.99±2.66 | 55.60±2.76 | 44.69±3.58 | 50.36±3.20 | 64.22±3.04 | |
|---|---|---|---|---|---|---|
| IPD, d=12 | 61.74±4.07 | 60.67±2.96 | 47.04±4.19 | 52.51±3.72 | 66.49±3.33 | 0.6653±0.0059 |
| | 61.61±3.60 | 59.25±2.44 | 48.43±3.59 | 53.81±3.17 | 66.87±2.86 | |
| $S_{2/3}$-LRR [68] | 54.98±3.75 | 51.54±3.21 | 36.46±3.80 | 42.87±3.42 | 58.87±3.47 | 0.4982±0.0078 |
| | 56.65±3.20 | 53.47±2.35 | 40.81±3.17 | 46.91±2.84 | 61.27±2.89 | |
| IPD, d=9 | 56.84±3.99 | 55.16±2.85 | 40.07±3.62 | 46.23±3.22 | 60.94±3.07 | 0.4927±0.0073 |
| | 57.65±3.99 | 55.62±2.48 | 42.91±3.65 | 48.89±3.22 | 62.51±3.05 | |
| WP $S_{2/3}$-LRR | 60.05±4.10 | 57.28±3.42 | 43.32±4.44 | 48.98±3.98 | 63.89±3.56 | 0.6734±0.0076 |
| | 60.26±3.64 | 56.71±2.98 | 45.46±3.92 | 51.08±3.50 | 64.71±3.16 | |
| IPD, d=10 | 63.66±4.60 | 64.04±2.91 | 49.53±4.89 | 54.79±3.89 | 68.22±3.49 | 0.6613±0.0067 |
| | 62.85±4.06 | 61.13±2.69 | 49.68±3.91 | 54.96±3.45 | 67.66±2.90 | |
| | **Deep** | **networks** | | | | |
| DSLSP [69] | <u>91.64</u> | <u>82.96</u> | - | - | - | - |
| AASSC [19] | 84.90 | 76.09 | - | - | 84.60 | |
| DSC-L2 [18], reported from [19] | 75.00 | 73.19 | - | - | 79.91 | - |
| DASC [21] | 80.40 | 78.00 | - | - | 83.70 | - |
| PSSC [22] | 84.30 | 76.76 | - | - | 84.30 | - |
| MESC [18] | 81.11 | 82.26 | - | - | - | - |

TABLE V: Clustering results on USPS dataset. All WPD SC results were obtained in A sub-band.

| Algorithm | ACC [%] in-sample out-of-sample | NMI [%] in-sample out-of-sample | Rand [%] in-sample out-of-sample | F_score[%] in-sample out-of-sample | Purity [%] in-sample out-of-sample | Average affinity |
|---|---|---|---|---|---|---|
| WP MERA [27] | **99.70±0.26** | **99.39±0.52** | **99.34±0.57** | **99.41±0.51** | **99.70±0.26** | |
| | **92.12±1.19** | **86.00±1.84** | **83.45±2.34** | **85.08±2.11** | **92.12±1.19** | |
| SSC [9] | 75.11±5.72 | 73.74±3.07 | 62.69±5.20 | 66.64±4.53 | 76.67±4.86 | 0.6358±0.0073 |
| | 78.19±5.03 | 74.02±2.39 | 70.19±4.56 | 73.62±3.97 | 79.75±4.05 | |
| IPD [35], d=11 | 75.45±5.68 | 74.95±3.24 | 64.19±5.51 | 68.01±4.81 | 77.21±4.85 | 0.6356±0.0065 |
| | 78.69±4.57 | 75.26±2.39 | 71.75±4.17 | 75.02±3.63 | 80.34±3.87 | |
| WP SSC | 75.39±5.78 | 73.75±2.85 | 62.84±4.93 | 66.64±4.30 | 76.98±4.77 | 0.6726±0.0074 |
| | 78.23±5.09 | 73.86±2.48 | 70.19±4.64 | 73.61±4.06 | 79.87±3.89 | |
| GMC LRSSC [10] | 79.47±4.76 | 76.91±2.46 | 68.00±4.00 | 71.25±3.56 | 81.95±3.11 | 0.6414±0.0050 |
| | 79.24±5.15 | 75.70±1.99 | 70.57±4.29 | 73.75±3.85 | 83.49±2.17 | |
| IPD, d=12 | 78.44±5.18 | 78.37±2.65 | 68.63±4.49 | 71.86±3.98 | 81.70±3.52 | 0.6360±0.0045 |
| | 79.98±4.68 | 77.39±1.94 | 73.19±4.12 | 76.13±3.69 | 84.17±2.45 | |
| WP GMC LRSSC | 80.10±5.18 | 76.46±2.75 | 67.93±4.48 | 71.16±3.99 | 82.00±3.56 | 0.6781±0.0051 |
| | 80.03±5.26 | 75.35±1.95 | 70.78±4.37 | 73.94±3.23 | 83.41±2.31 | |
| IPD, d=12 | 77.87±5.31 | 78.42±2.67 | 68.46±4.42 | 71.72±4.11 | 81.36±3.65 | 0.6734±0.0049 |
| | 79.28±5.19 | 77.29±2.10 | 72.90±4.48 | 75.86±4.01 | 83.96±2.59 | |
| S0L0 LRSSC [10] | 82.75±6.14 | 80.57±3.09 | 72.69±5.65 | 75.47±5.03 | 84.37±4.53 | 0.6376±0.0056 |
| | 83.64±5.60 | 78.95±2.34 | 76.56±5.29 | 79.13±4.73 | 85.93±3.14 | |
| WP S0L0 LRSSC | 83.06±5.40 | 80.38±2.67 | 72.69±4.82 | 75.47±4.29 | 84.58±3.90 | 0.6752±0.0051 |
| | 83.68±5.43 | 78.88±2.15 | 76.66±4.88 | 79.23±4.35 | 85.85±2.99 | |
| LRR [8] | 69.24±3.66 | 68.90±2.78 | 50.20±5.30 | 55.84±4.54 | 72.59±3.04 | 0.6411±0.0060 |
| | 72.25±3.47 | 70.84±2.13 | 58.75±4.19 | 63.51±3.58 | 77.77±2.21 | |
| IPD, d=12 | 80.77±5.67 | 80.76±2.41 | 71.09±4.27 | 74.18±4.34 | 82.08±4.59 | 0.6628±0.0344 |
| | 82.02±5.50 | 79.18±2.10 | 75.74±5.08 | 78.49±4.51 | 83.97±3.65 | |
| WP LRR | 69.58±4.17 | 69.40±2.65 | 50.99±4.87 | 56.43±4.20 | 72.97±3.26 | 0.6763±0.0060 |
| | 72.50±3.90 | 71.20±2.10 | 59.42±3.82 | 64.08±3.31 | 77.92±2.37 | |
| IPD, d=12 | 79.87±5.36 | 80.57±2.42 | 70.29±4.85 | 73.50±4.23 | 80.69±4.71 | 0.7108±0.0360 |
| | 81.68±5.01 | 78.92±2.08 | 75.66±4.49 | 78.46±3.96 | 82.85±3.67 | |
| NSN [67] | 74.67±5.40 | 70.24±3.50 | 58.32±5.20 | 63.04±4.83 | 76.75±4.07 | 0.6511±0.0055 |
| | 77.13±5.84 | 72.31±2.81 | 66.14±5.57 | 70.05±5.25 | 80.50±3.25 | |
| IPD, d=12 | 75.72±5.44 | 70.44±3.99 | 59.66±5.86 | 63.82±5.21 | 77.31±4.31 | 0.6510±0.0053 |
| | 78.46±5.89 | 72.67±3.25 | 67.47±6.12 | 71.13±5.41 | 80.99±3.71 | |
| WP NSN | 74.39±5.32 | 69.83±3.37 | 58.59±5.79 | 62.66±4.59 | 76.30±4.04 | 0.6870±0.0055 |
| | 77.08±5.62 | 71.95±2.72 | 66.35±5.82 | 69.96±4.94 | 80.14±3.16 | |
| IP, d=12 | 74.91±5.16 | 70.26±3.63 | 58.99±5.29 | 63.25±4.69 | 76.71±3.84 | 0.6861±0.0063 |
| | 77.88±5.68 | 72.40±2.88 | 66.93±5.94 | 70.66±5.27 | 80.57±3.25 | |
| RTSC [11] | 72.11±5.97 | 69.54±3.63 | 58.08±5.41 | 62.34±4.82 | 75.31±4.20 | 0.6557±0.0056 |
| | 75.37±5.46 | 71.50±2.73 | 65.20±5.39 | 68.97±4.83 | 79.09±3.26 | |

| Algorithm | ACC | NMI | Rand | F_score | Purity | Average affinity |
|---|---|---|---|---|---|---|
| IPD, d=12 | 72.35±5.72 | 70.31±3.34 | 58.78±5.24 | 62.97±4.68 | 75.57±4.23 | 0.6437±0.0053 |
| | 75.10±5.32 | 71.84±2.42 | 65.27±4.90 | 69.03±4.39 | 79.12±3.21 | |
| WP RTSC | 71.65±5.42 | 69.72±3.47 | 58.19±5.17 | 62.45±4.61 | 75.14±3.97 | 0.6807±0.0055 |
| | 75.36±5.13 | 71.63±2.71 | 65.69±5.16 | 69.41±4.63 | 79.10±2.99 | |
| IPD, d=12 | 71.42±5.00 | 70.02±3.13 | 58.34±4.75 | 62.58±4.24 | 75.09±3.72 | 0.6804±0.0052 |
| | 74.91±4.82 | 71.64±2.43 | 65.49±4.85 | 69.24±4.36 | 78.90±2.83 | |
| $S_{1/2}$-LRR [68] | 77.24±3.49 | 69.18±2.54 | 59.61±3.67 | 63.63±3.29 | 77.63±2.91 | 0.6531±0.0056 |
| | 79.06±3.33 | 70.94±1.86 | 66.14±3.59 | 69.81±3.22 | 79.86±2.21 | |
| IPD, d=12 | 78.04±6.00 | 73.60±3.26 | 64.14±5.67 | 67.79±5.03 | 79.79±4.26 | 0.6430±0.0056 |
| | 80.54±5.84 | 74.94±2.57 | 71.57±5.55 | 74.69±4.96 | 83.06±3.23 | |
| WP $S_{1/2}$-LRR | 77.32±3.26 | 69.18±2.47 | 59.65±3.54 | 63.66±3.18 | 77.63±2.82 | 0.6892±0.0051 |
| | 78.88±3.08 | 70.45±1.89 | 65.64±3.49 | 69.37±3.12 | 79.52±2.23 | |
| IPD, d=12 | 76.70±6.34 | 73.00±3.50 | 63.00±5.91 | 66.78±5.22 | 78.87±4.44 | 0.6804±0.0046 |
| | 78.95±6.17 | 74.02±2.65 | 70.10±5.75 | 73.39±5.13 | 82.15±3.23 | |
| $S_{2/3}$-LRR [68] | 75.78±3.92 | 68.67±2.53 | 58.40±3.59 | 62.55±3.22 | 76.60±2.99 | 0.6539±0.0053 |
| | 77.70±3.86 | 70.47±1.71 | 65.07±3.42 | 68.86±3.06 | 79.28±2.09 | |
| IPD, d=14 | 78.98±5.47 | 74.38±2.99 | 65.31±5.14 | 68.82±4.58 | 80.58±3.86 | 0.6429±0.0053 |
| | 80.96±5.38 | 75.16±2.37 | 71.75±5.19 | 74.83±4.65 | 83.26±2.95 | |
| WP $S_{2/3}$-LRR | 75.61±3.96 | 68.58±2.72 | 58.22±3.72 | 62.38±3.33 | 76.43±3.04 | 0.6893±0.0047 |
| | 77.28±3.59 | 69.95±1.69 | 64.30±3.26 | 68.19±2.92 | 78.78±2.08 | |
| IPD, d=14 | 78.61±5.68 | 74.31±2.85 | 65.23±4.89 | 68.75±4.35 | 80.43±3.86 | 0.6791±0.0049 |
| | 80.12±5.21 | 74.63±2.38 | 70.96±5.41 | 74.13±4.85 | 82.98±2.92 | |
| **Deep networks** | | | | | | |
| DSLSP [69] | <u>83.29</u> | <u>83.70</u> | - | - | - | - |
| DSC-L2 [17], reported from [19] | 77.64 | 78.86 | - | - | - | - |
| MESC [18] | 81.49 | 86.34 | - | - | - | - |

TABLE VI: Clustering results on EYaleB dataset. Sub-bands for each WPD SC algorithm are reported in table.

| Algorithm | ACC [%] | NMI [%] | Rand [%] | F_score[%] | Purity [%] | Average affinity |
|---|---|---|---|---|---|---|
| | in-sample out-of-sample | **in-sample** out-of- | **in-sample** out-of- | **in-sample** out-of- | **in-sample** out-of- | |

| | | sample | sample | sample | sample | |
|---|---|---|---|---|---|---|
| WP MERA [27] | **99.49±1.31** | **99.89±0.28** | **99.51±1.25** | **99.52±1.21** | **99.60±1.02** | |
| | **92.93±1.39** | **92.83±1.03** | **86.17±2.07** | **86.52±2.02** | **93.00±1.26** | |
| SSC [9] | 75.65±1.84 | 80.43±1.65 | 38.59±4.66 | 40.77±4.39 | 76.29±1.67 | 0.5890±0.0038 |
| | 81.36±2.17 | 85.79±1.67 | 57.45±3.97 | 58.73±3.81 | 82.01±2.05 | |
| IPD [35], d=3 | 87.64±2.14 | 91.32±0.78 | 80.50±2.11 | 81.02±2.05 | 88.12±1.92 | 0.6006±0.0065 |
| | 86.54±2.18 | 90.49±1.05 | 77.46±2.91 | 78.06±2.82 | 86.96±2.04 | |
| WP-D SSC | 95.01±2.58 | 98.05±0.72 | 94.10±2.70 | 94.26±2.63 | 95.92±2.01 | 0.1996±0.0022 |
| | 91.49±2.55 | 94.05±0.92 | 86.71±2.76 | 87.06±2.68 | 92.26±2.06 | |
| IPD, d=8 | <u>96.92±1.68</u> | <u>98.63±0.42</u> | <u>96.24±1.60</u> | <u>96.34±1.55</u> | <u>97.46±1.27</u> | <u>0.2007±0.0019</u> |
| | <u>93.07±1.62</u> | <u>94.45±0.71</u> | <u>88.38±1.71</u> | <u>88.67±1.66</u> | <u>93.57±1.24</u> | |
| GMC LRSSC [10] | 88.69±1.73 | 91.10±1.11 | 78.36±3.00 | 78.95±2.91 | 89.96±1.63 | 0.5943±0.0073 |
| | 87.06±1.67 | 89.83±1.13 | 78.49±2.77 | 79.05±2.69 | 87.37±1.54 | |
| IPD, d=8 | 89.27±1.21 | 91.82±0.68 | 81.21±1.95 | 81.72±1.89 | 89.62±1.53 | 0.6080±0.0050 |
| | 88.65±1.90 | 91.47±0.89 | 80.02±2.30 | 80.54±2.23 | 89.01±1.74 | |
| WP-DH GMC LRSSC | 87.74±1.62 | 90.85±0.73 | 76.98±2.60 | 77.62±2.52 | 88.31±1.32 | 0.2125±0.0013 |
| | 87.95±1.90 | 91.30±0.99 | 77.48±3.01 | 78.09±2.92 | 88.55±1.63 | |
| IPD, d=9 | 88.25±1.97 | 90.96±0.78 | 77.79±2.57 | 78.40±2.49 | 89.16±1.28 | 0.2129±0.0009 |
| | 88.49±1.82 | 91.50±0.88 | 78.27±2.83 | 78.85±2.75 | 89.90±1.67 | |
| S0L0 LRSSC [10] | 87.98±1.78 | 90.88±0.85 | 77.77±2.62 | 78.38±2.53 | 88.36±1.61 | 0.6052±0.0066 |
| | 87.83±1.81 | 90.97±1.12 | 78.15±2.51 | 78.73±2.44 | 88.23±1.69 | |
| WP-DA S0L0 LRSSC | 81.82±2.90 | 87.51±1.11 | 64.87±3.90 | 65.92±3.76 | 82.92±2.30 | 0.2390±0.0026 |
| | 81.75±2.91 | 87.63±1.30 | 65.93±3.88 | 66.91±3.74 | 82.79±2.56 | |
| IPD, d=6 | 82.88±2.48 | 88.86±1.01 | 72.49±2.90 | 73.26±2.81 | 83.88±2.10 | 0.2396±0.0026 |
| | 81.85±2.72 | 88.13±1.23 | 69.42±3.60 | 70.27±3.49 | 82.82±2.31 | |
| LRR [8] | 73.38±2.59 | 83.10±1.34 | 46.31±5.99 | 48.22±5.67 | 74.21±2.40 | 0.5689±0.0089 |
| | 76.24±2.56 | 85.93±1.31 | 56.74±5.34 | 58.14±5.10 | 76.77±2.49 | |
| IPD, d=9 | 81.45±2.87 | 88.17±1.00 | 61.36±5.22 | 62.59±5.00 | 82.16±2.60 | 0.5845±0.0102 |
| | 83.07±2.93 | 89.56±1.15 | 68.77±4.48 | 69.69±4.31 | 83.55±2.78 | |
| WP-D LRR | 66.31±3.17 | 76.47±1.55 | 22.49±3.82 | 25.74±3.56 | 68.88±2.50 | 0.1805±0.0035 |
| | 74.49±3.08 | 83.40±1.43 | 43.56±5.25 | 45.55±4.98 | 77.18±2.39 | |
| IPD, d=9 | 79.21±2.63 | 85.36±1.23 | 47.64±6.12 | 49.48±5.81 | 80.36±2.32 | 0.1873±0.0034 |
| | 82.90±2.50 | 88.37±1.12 | 62.58±4.44 | 63.73±4.26 | 84.00±2.10 | |

| | | | | | |
|---|---|---|---|---|---|
| NSN [67] | 72.98±2.53 | 75.35±1.46 | 56.65±2.58 | 57.72±2.51 | 73.48±2.39 | 0.5689±0.0072 |
| | 74.86±2.12 | 79.81±1.16 | 60.07±2.23 | 61.12±2.17 | 75.40±2.04 | |
| WP-DH NSN | 88.08±1.66 | 89.75±0.75 | 80.19±1.80 | 80.71±1.75 | 88.28±1.57 | 0.2122±0.0011 |
| | 87.32±1.73 | 89.78±0.94 | 77.76±2.30 | 78.34±2.23 | 87.61±1.65 | |
| RTSC [11] | 40.50±1.62 | 52.26±1.17 | 17.83±1.51 | 20.53±1.38 | 42.56±1.36 | 0.4889±0.0055 |
| | 46.25±2.06 | 59.92±1.29 | 24.73±2.09 | 27.00±1.97 | 47.96±1.88 | |
| WP-DH RTSC | 87.71±1.38 | 89.81±0.77 | 74.29±2.46 | 75.00±2.38 | 87.97±1.15 | 0.2118±0.0009 |
| | 88.14±1.54 | 90.75±0.89 | 76.26±2.78 | 76.89±2.70 | 88.54±1.34 | |
| $S_{1/2}$-LRR [68] | 68.44±1.73 | 74.80±1.12 | 52.80±1.69 | 53.83±1.63 | 68.93±1.67 | 0.5661±0.0070 |
| | 70.89±2.19 | 79.24±1.49 | 57.33±3.13 | 58.49±3.03 | 71.51±2.10 | |
| IPD, d=9 | 91.20±1.10 | 92.82±0.49 | 83.59±1.57 | 84.03±1.52 | 91.37±1.01 | 0.6128±0.0036 |
| | 90.43±1.38 | 92.40±0.90 | 82.31±2.45 | 82.77±2.38 | 90.68±1.28 | |
| WP-DA $S_{1/2}$-LRR | 70.45±1.85 | 75.80±1.28 | 49.52±2.64 | 51.01±2.52 | 71.05±1.72 | 0.2324±0.0019 |
| | 73.46±2.08 | 80.01±1.41 | 55.42±3.16 | 56.68±3.05 | 74.17±2.00 | |
| IPD, d=8 | 87.21±1.59 | 90.16±0.70 | 75.79±2.09 | 76.46±2.02 | 87.60±1.38 | 0.2430±0.0018 |
| | 86.52±1.80 | 89.84±0.97 | 73.73±2.89 | 74.44±2.80 | 86.92±1.63 | |
| $S_{2/3}$-LRR [68] | 68.54±1.98 | 74.90±1.19 | 52.50±1.86 | 53.84±1.79 | 69.03±1.89 | 0.5659±0.0091 |
| | 70.94±2.35 | 79.43±1.29 | 57.43±2.82 | 58.59±2.72 | 71.55±2.15 | |
| IPD, d=9 | 91.56±1.55 | 92.93±0.61 | 83.82±1.77 | 84.25±1.71 | 91.71±1.43 | 0.6124±0.0033 |
| | 90.81±1.58 | 92.56±0.85 | 82.56±2.37 | 83.01±2.31 | 91.01±1.48 | |
| WP-D $S_{2/3}$-LRR | 69.59±1.94 | 75.10±1.31 | 48.65±2.45 | 50.16±2.34 | 70.28±1.34 | 0.2317±0.0020 |
| | 72.98±2.06 | 79.67±1.36 | 54.86±3.40 | 56.13±2.37 | 73.74±1.87 | |
| IPD, d=7 | 88.12±1.70 | 90.89±0.76 | 77.78±2.48 | 78.39±2.40 | 88.51±1.41 | 0.1939±0.0012 |
| | 87.87±1.95 | 91.09±1.01 | 77.42±2.85 | 78.02±2.77 | 88.37±1.64 | |
| **Deep networks** | | | | | | |
| TSC LLMC [12] | 98 | 85 | 91 | 90 | 95 | |
| TSC SSC [12] | <u>99</u> | <u>94</u> | <u>96</u> | <u>98</u> | <u>98</u> | |
| TSC LRR [12] | 69 | 74 | 75 | 74 | 72 | |
| DSLSP [69] | 97.62 | 96.74 | - | - | - | - |
| DSC-L2 [17], reported | 97.73 | 97.03 | - | - | - | - |

| | | | | | |
|---|---|---|---|---|---|
| from [19] | | | | | |
| DASC [21] | <u>98.56</u> | <u>98.01</u> | - | - | - | - |
| MESC [18] | 98.03 | 97.27 | - | - | - | - |
| SAE [33] | 88.75 | 87.53 | - | - | - | - |
| DCSC [24] | 92.36 | 94.27 | - | - | - | - |
| LBDR [25] | 84.73 | 86.75 | - | - | - | - |

TABLE VII: Clustering results on ORL dataset. Sub-bands for each WPD SC algorithm are reported in table.

| Algorithm | ACC [%] in-sample out-of-sample | NMI [%] in-sample out-of-sample | Rand [%] in-sample out-of-sample | F_score[%] in-sample out-of-sample | Purity [%] in-sample out-of-sample | Average affinity |
|---|---|---|---|---|---|---|
| WP MERA [27] | 81.71±2.83 80.31±3.36 | 91.28±1.36 92.31±1.34 | 73.47±3.64 66.02±5.02 | 74.08±3.56 66.68±4.91 | 84.08±2.47 81.98±3.06 | |
| IPD [35], d=5 | <u>88.98±2.69</u> <u>86.72±2.98</u> | <u>94.02±1.27</u> <u>94.41±1.26</u> | <u>81.95±3.67</u> <u>74.95±4.94</u> | <u>82.35±3.59</u> <u>75.41±4.84</u> | <u>89.98±2.30</u> <u>87.56±2.80</u> | |
| SSC [9] | 72.23±2.93 70.06±3.27 | 86.35±1.32 87.46±1.57 | 59.66±3.57 48.74±5.46 | 60.61±3.48 49.82±5.31 | 75.27±2.53 72.13±2.96 | 0.5220±0.0137 |
| IPD, d=7 | 75.31±3.09 72.16±3.13 | 88.04±1.38 88.39±1.40 | 64.07±3.77 51.80±4.71 | 64.91±3.67 52.80±4.58 | 78.45±2.44 74.38±2.87 | 0.5090±0.0117 |
| WP-AH SSC | 73.72±2.66 69.71±3.12 | 86.87±1.23 87.02±1.39 | 61.67±3.20 47.84±4.55 | 62.56±3.12 48.92±4.42 | 76.64±2.24 71.93±2.92 | 0.4993±0.0051 |
| IPD, d=4 | 76.14±3.07 72.21±3.00 | 88.63±1.50 88.18±1.50 | 65.78±4.05 51.41±5.09 | 65.58±3.95 52.41±4.96 | 79.47±2.73 74.57±2.85 | 0.4146±0.0055 |
| GMC LRSSC [10] | 77.97±2.10 75.14±2.87 | 88.45±1.29 89.28±1.39 | 67.12±3.35 55.86±5.14 | 67.81±3.27 56.73±5.01 | 79.84±1.21 76.74±2.72 | 0.3281±0.0052 |
| IPD, d=5 | 81.23±2.64 77.22±3.05 | 90.08±1.17 90.29±1.27 | 71.25±3.16 59.37±4.58 | 71.90±3.08 60.16±4.47 | 82.90±2.15 78.74±2.77 | 0.5204±0.0100 |
| WP-A GMC LRSSC | 78.61±2.38 76.07±2.81 | 89.00±1.18 89.94±1.20 | 68.09±3.09 58.02±4.30 | 68.81±3.02 58.85±4.20 | 80.59±2.11 77.73±2.54 | 0.5370±0.0098 |
| IPD, d=4 | 81.01±2.47 77.64±2.70 | 90.30±1.18 90.71±1.36 | 71.37±2.34 60.99±4.84 | 72.02±3.16 61.75±4.73 | 82.89±2.13 79.26±2.62 | 0.5822±0.0092 |
| S0L0 LRSSC [10] | 63.81±2.82 | 80.35±1.40 | 48.40±3.17 | 49.58±3.09 | 67.03±2.36 | 0.5282±0.0101 |
| | 65.39±3.34 | 84.86±1.55 | 42.72±4.67 | 43.88±4.55 | 67.90±3.16 | |
| IPD d=5 | 66.48±2.05 | 81.25±1.08 | 50.56±2.44 | 51.69±2.38 | 68.83±1.85 | 0.5058±0.0114 |
| | 65.67±3.02 | 84.81±1.42 | 42.50±4.32 | 43.68±4.20 | 67.81±2.90 | |
| WP-AH S0L0 LRSSC | 74.73±2.51 | 86.30±1.27 | 61.49±3.36 | 62.37±3.28 | 77.26±2.22 | 0.3855±0.0060 |
| | 72.47±3.35 | 88.06±1.44 | 51.99±4.87 | 52.95±4.76 | 74.38±2.76 | |
| IPD, d=5 | 76.74±2.72 | 87.25±1.21 | 64.21±3.30 | 65.03±3.22 | 79.04±2.20 | 0.3886±0.0067 |
| | 74.10±2.99 | 88.58±1.43 | 53.93±4.64 | 54.85±4.53 | 75.62±2.71 | |
| LRR [8] | 66.88±2.55 | 82.70±1.26 | 46.50±4.35 | 47.91±4.17 | 71.31±1.98 | 0.5383±0.0142 |
| | 65.08±3.24 | 84.18±1.72 | 35.73±5.99 | 37.25±5.76 | 68.08±2.94 | |
| IPD d=5 | 70.33±2.83 | 85.09±1.29 | 55.40±3.90 | 56.53±3.77 | 72.80±2.44 | 0.5908±0.0249 |
| | 67.90±3.02 | 86.12±1.33 | 44.51±4.65 | 45.73±4.49 | 69.49±2.93 | |
| WP-AA LRR | 68.21±3.10 | 83.95±1.58 | 47.82±5.92 | 49.23±5.68 | 72.71±2.48 | 0.6279±0.0172 |
| | 65.10±3.97 | 84.21±2.32 | 35.80±7.08 | 37.34±6.82 | 67.75±3.81 | |
| IPD, d=6 | 72.14±2.99 | 87.06±1.37 | 58.95±4.46 | 59.99±4.31 | 74.86±2.72 | 0.6718±0.0278 |
| | 69.81±2.92 | 87.61±1.41 | 48.63±5.10 | 49.74±4.95 | 71.53±2.75 | |
| NSN [67] | 67.80±2.52 | 82.78±1.27 | 51.98±3.21 | 53.11±3.12 | 70.49±2.13 | 0.3187±0.0055 |
| | 65.05±2.86 | 84.65±1.29 | 41.18±3.98 | 42.43±3.87 | 67.33±2.79 | |
| WP-AH NSN | 69.26±2.64 | 83.77±1.33 | 55.27±3.18 | 56.34±3.09 | 72.05±2.30 | 0.4048±0.0080 |
| | 67.92±3.36 | 85.88±1.59 | 45.59±4.94 | 46.73±4.81 | 69.76±3.25 | |
| RTSC [11] | 69.12±2.70 | 82.86±1.40 | 53.31±3.43 | 54.38±3.34 | 71.65±2.33 | 0.4870±0.0065 |
| | 66.57±2.89 | 85.27±1.40 | 43.30±4.35 | 44.48±4.23 | 68.77±2.88 | |
| WP-AH RTSC | 69.99±2.75 | 82.84±1.42 | 54.00±3.52 | 55.05±3.43 | 72.45±2.50 | 0.3798±0.0039 |
| | 69.47±2.33 | 86.35±1.33 | 46.95±4.11 | 48.02±4.01 | 71.50±2.25 | |
| $S_{1/2}$-LRR [68] | 67.36±2.78 | 82.47±1.41 | 52.41±3.23 | 53.51±3.15 | 70.55±2.46 | 0.3331±0.0053 |
| | 66.92±3.09 | 85.53±1.51 | 44.23±4.68 | 45.37±4.55 | 69.18±2.88 | |
| IPD, d=6 | 74.68±2.49 | 86.58±1.20 | 62.34±3.05 | 63.20±2.97 | 77.19±2.00 | 0.5082±0.0121 |
| | 72.38±2.82 | 88.22±1.22 | 52.48±4.35 | 53.43±4.24 | 74.31±2.64 | |
| WP-AA $S_{1/2}$-LRR | 68.24±2.78 | 83.09±1.31 | 54.00±3.17 | 50.05±3.09 | 71.28±2.35 | 0.5185±0.0069 |
| | 66.81±3.19 | 85.57±1.54 | 45.38±4.55 | 45.73±4.48 | 69.29±3.03 | |
| IPD, d=6 | 77.18±2.43 | 87.80±1.15 | 65.45±3.00 | 66.23±2.93 | 79.28±2.07 | 0.5989±0.0114 |
| | 74.83±3.18 | 89.18±1.47 | 55.98±5.22 | 56.84±5.10 | 76.37±3.07 | |
| $S_{2/3}$-LRR [68] | 68.00±3.00 | 82.88±1.47 | 53.57±3.53 | 54.63±3.45 | 70.99±2.63 | 0.5093±0.0117 |
| | 67.33±3.31 | 85.90±1.43 | 45.42±4.63 | 46.53±4.51 | 69.72±3.02 | |
| IPD d=6 | 75.76±2.62 | 86.92±1.26 | 63.44±3.21 | 64.27±3.13 | 77.91±2.17 | 0.5054±0.0122 |
| | 73.57±2.86 | 88.61±1.21 | 53.69±4.43 | 54.62±4.32 | 75.33±2.69 | |
| WP-AH $S_{2/3}$-LRR | 68.86±2.82 | 83.18±1.49 | 54.54±3.44 | 55.58±3.36 | 71.86±2.55 | 0.4053±0.0062 |
| | 68.43±3.23 | 86.22±1.64 | 46.51±5.16 | 47.59±5.04 | 70.05±3.04 | |
| IPD d=5 | 81.55±2.59 | 90.07±1.23 | 71.47±3.24 | 72.11±3.16 | 83.25±2.25 | 0.3956±0.0063 |
| | 77.78±3.02 | 90.08±1.30 | 59.43±4.66 | 60.21±4.56 | 78.99±2.78 | |
| | **Deep** | **networks** | | | | |
| DSLSP [69] | 87.55 | 92.49 | - | - | - | - |
| AASSC [19] | **90.75** | **94.31** | - | - | 91.75 | |
| DSC-L2 [17], reported from [19] | 86.00 | 90.34 | - | - | - | - |
| DASC [21] | 88.25 | 93.15 | - | - | 89.25 | - |
| PSSC [22] | 86.75 | 93.49 | - | - | 89.25 | |
| MESC [18] | **90.25** | **93.59** | - | - | - | - |
| SAE [23] | 74.81 | 88.0 | - | - | - | |
| DCSC [24] | 83.52 | 90.1 | - | - | - | - |
| LBDR [25] | 77.68 | 89.12 | - | - | - | - |

TABLE VIII: Clustering results on COIL20 dataset. Sub-bands for each WPD SC algorithm are reported in table.

| Algorithm | ACC [%] in-sample out-of-sample | NMI [%] in-sample out-of-sample | Rand [%] in-sample out-of-sample | F_score[%] in-sample out-of-sample | Purity [%] in-sample out-of-sample | Average affinity |
|---|---|---|---|---|---|---|
| WP MERA [27] | 96.04±4.59 | 98.82±1.33 | 96.04±4.56 | 96.24±4.33 | 97.02±3.41 | |
| | 94.19±4.45 | 96.48±1.53 | 92.34±4.42 | 92.73±4.18 | 95.04±3.48 | |
| IPD [35], d=10 | **99.94±0.15** | **99.93±0.18** | **99.88±0.31** | **99.88±0.29** | **99.94±0.15** | |
| | **98.02±0.72** | **97.63±0.77** | **96.02±1.39** | **96.22±1.32** | **98.02±0.72** | |
| SSC [9] | 70.55±3.40 | 82.44±1.69 | 61.63±3.51 | 63.66±3.29 | 74.14±2.60 | 0.3838±0.0102 |
| | 68.99±3.40 | 81.10±1.61 | 69.37±3.35 | 62.54±3.12 | 72.28±3.60 | |
| IPD, d=8 | 75.93±2.88 | 85.71±1.45 | 68.49±3.44 | 70.12±3.23 | 78.98±2.16 | 0.3909±0.0063 |
| | 74.04±2.79 | 84.29±1.54 | 66.91±3.63 | 68.68±3.39 | 76.85±2.19 | |
| WP-A SSC | 72.03±3.31 | 82.95±1.79 | 63.27±3.94 | 65.18±3.69 | 75.10±2.77 | 0.4252±0.0092 |
| | 70.00±3.32 | 81.53±1.81 | 61.77±3.96 | 63.83±3.69 | 73.05±2.80 | |
| IPD, d=8 | 76.53±3.11 | 86.13±1.65 | 69.43±3.50 | 70.99±3.29 | 79.84±2.36 | 0.4293±0.0063 |
| | 74.71±2.92 | 84.75±1.87 | 67.96±3.53 | 69.66±3.32 | 77.75±2.46 | |
| GMC LRSSC [10] | 71.45±2.70 | 82.98±2.41 | 61.37±3.61 | 63.46±3.35 | 75.02±2.12 | 0.3733±0.0073 |
| | 69.93±2.39 | 81.70±1.44 | 60.04±3.64 | 62.28±3.36 | 73.06±2.00 | |
| IPD, d=6 | 71.71±2.58 | 83.01±1.52 | 61.64±3.51 | 63.71±3.27 | 75.11±2.06 | 0.4055±0.0076 |
| | 70.23±2.51 | 81.70±1.64 | 60.85±3.42 | 63.02±3.18 | 73.33±2.06 | |
| WP-AH GMC LRSSC | 75.59±2.44 | 83.30±1.37 | 67.12±2.72 | 68.76±2.57 | 77.83±1.99 | 0.3159±0.0039 |
| | 74.44±2.10 | 82.02±1.33 | 66.41±2.52 | 68.14±2.37 | 76.44±1.72 | |
| IPD, d=6 | 76.19±2.33 | 83.80±1.36 | 67.63±2.60 | 69.26±2.45 | 78.40±1.86 | 0.2928±0.0049 |
| | 74.86±1.97 | 82.36±1.32 | 66.83±2.44 | 68.54±2.29 | 77.03±1.57 | |
| S0L0 LRSSC [10] | 69.93±2.99 | 81.14±1.54 | 60.41±3.10 | 62.48±2.90 | 73.28±2.35 | 0.3839±0.0085 |
| | 68.62±2.74 | 79.89±1.63 | 59.59±3.07 | 61.76±2.87 | 71.68±2.31 | |
| IPD, d=8 | 71.65±2.80 | 81.99±1.49 | 62.28±3.11 | 64.23±2.92 | 74.63±2.28 | 0.3930±0.0080 |
| | 70.18±2.72 | 80.64±1.55 | 61.31±3.13 | 63.37±2.93 | 72.96±2.29 | |
| WP-HA S0L0 LRSSC | 71.87±2.66 | 81.18±1.60 | 61.59±3.39 | 63.60±3.16 | 74.93±2.07 | 0.2942±0.0051 |
| | 70.45±2.41 | 79.63±1.43 | 60.78±3.11 | 62.89±2.88 | 73.31±1.83 | |
| LRR [8] | 61.30±4.42 | 75.57±2.01 | 49.77±2.09 | 52.58±4.66 | 64.77±3.55 | 0.3771±0.0079 |
| | 59.32±4.34 | 74.48±2.04 | 49.12±2.03 | 52.07±4.59 | 62.33±3.65 | |
| IPD, d=7 | 70.85±3.87 | 82.97±1.73 | 60.59±4.86 | 62.73±4.50 | 74.66±2.88 | 0.3790±0.0079 |
| | 69.34±3.62 | 81.46±1.70 | 59.39±4.45 | 61.67±4.11 | 72.65±2.70 | |
| WP-AH LRR | 70.76±3.00 | 80.68±1.44 | 61.55±3.06 | 63.55±2.86 | 73.90±3.49 | 0.3173±0.0048 |
| | 69.82±3.04 | 79.86±1.43 | 61.90±3.00 | 63.93±2.80 | 72.80±2.62 | |
| IPD, d=10 | 72.90±3.10 | 81.63±1.55 | 62.14±3.29 | 64.14±3.93 | 75.79±2.23 | 0.3270±0.0077 |
| | 71.65±2.87 | 80.36±1.73 | 61.60±3.03 | 63.67±2.84 | 74.26±2.19 | |
| NSN [67] | 74.02±3.24 | 83.50±1.60 | 65.93±3.27 | 67.69±3.08 | 76.30±2.85 | 0.4228±0.0128 |
| | 72.17±3.28 | 81.74±1.74 | 64.57±3.20 | 66.44±3.01 | 74.32±2.82 | |
| IPD, d=10 | 78.33±2.34 | 84.77±1.35 | 69.69±2.86 | 71.21±2.70 | 79.78±2.01 | 0.3767±0.0058 |
| | 76.08±2.30 | 83.01±1.35 | 67.99±2.72 | 69.54±2.56 | 77.54±1.92 | |
| WP-AH NSN | 75.53±3.80 | 85.09±1.83 | 68.64±3.89 | 70.25±3.67 | 78.00±3.13 | 0.3026±0.0065 |
| | 73.42±3.64 | 83.09±1.85 | 66.92±3.58 | 68.66±3.37 | 76.01±3.02 | |
| IPD, d=10 | 81.56±2.40 | 86.88±1.46 | 74.16±2.81 | 75.43±2.66 | 82.74±2.16 | |
| | 79.34±2.10 | 85.22±1.42 | 72.29±2.45 | 73.69±2.32 | 80.58±1.95 | |
| RTSC [11] | 72.51±3.29 | 82.23±1.49 | 63.14±3.42 | 65.04±3.21 | 75.55±2.43 | 0.3778±0.0062 |
| | 70.79±2.89 | 80.71±1.43 | 61.89±3.07 | 63.93±2.87 | 73.55±2.18 | |
| IPD, d=5 | 73.94±3.23 | 82.92±1.54 | 65.07±3.36 | 66.86±3.16 | 76.83±2.48 | 0.4295±0.0074 |
| | 72.16±2.87 | 81.13±1.45 | 63.81±3.01 | 65.71±2.82 | 74.75±2.12 | |
| WP-AH RTSC | 74.84±3.15 | 83.89±1.42 | 65.36±3.83 | 67.16±3.58 | 78.20±2.15 | 0.3123±0.0047 |
| | 73.64±2.91 | 82.51±1.44 | 64.56±3.56 | 66.45±3.32 | 76.72±2.01 | |
| IPD, d=6 | 76.25±3.21 | 84.82±1.47 | 67.37±3.66 | 69.05±3.43 | 79.17±2.31 | 0.2910±0.0039 |
| | 75.17±2.92 | 83.44±1.52 | 66.84±3.26 | 68.59±3.05 | 77.84±2.22 | |
| $S_{1/2}$-LRR [68] | 64.85±2.79 | 75.57±1.43 | 54.58±2.57 | 56.83±2.44 | 66.61±2.32 | 0.3950±0.0060 |
| | 62.70±2.31 | 74.19±1.38 | 53.79±2.18 | 56.15±2.06 | 64.30±1.94 | |
| IPD, d=7 | 67.10±2.87 | 78.42±1.54 | 56.60±3.30 | 58.84±3.09 | 70.41±2.35 | 0.3946±0.0068 |
| | 65.40±2.75 | 77.11±1.67 | 55.99±3.15 | 58.32±2.94 | 68.23±2.40 | |
| WP-AH $S_{1/2}$-LRR | 72.00±2.84 | 79.52±1.54 | 63.11±2.80 | 64.92±2.66 | 73.68±2.28 | 0.3293±0.0039 |
| | 70.84±2.64 | 78.60±1.43 | 62.92±2.42 | 64.80±2.29 | 72.51±2.08 | |
| $S_{2/3}$-LRR [68] | 64.98±3.00 | 74.51±1.83 | 53.03±3.10 | 55.39±2.94 | 67.13±2.53 | 0.3940±0.0081 |
| | 63.13±2.67 | 73.29±1.69 | 52.56±2.60 | 55.03±2.46 | 65.06±2.31 | |
| IPD, d=8 | 69.08±2.79 | 79.46±1.47 | 58.31±3.20 | 60.46±2.99 | 72.05±2.26 | 0.3931±0.0088 |
| | 67.31±2.59 | 78.19±1.41 | 57.42±3.01 | 59.69±2.80 | 70.05±2.05 | |
| WP-AH $S_{2/3}$-LRR | 71.88±2.80 | 79.55±1.45 | 63.03±2.72 | 64.85±2.58 | 73.62±2.17 | 0.3296±0.0040 |
| | 70.58±2.76 | 78.58±1.51 | 62.85±2.67 | 64.72±2.53 | 72.33±2.17 | |
| | | **Deep** | **networks** | | | |
| TSC LLMC [12] | <u>98</u> | <u>94</u> | <u>94</u> | <u>97</u> | <u>99</u> | |
| TSC SSC [12] | 97 | 90 | 92 | 96 | 97 | |
| TSC LRR [12] | 78 | 83 | 72 | 74 | 72 | |
| DSLSP [69] | 97.57 | 97.40 | - | - | - | - |
| AASSC [19] | <u>98.40</u> | <u>98.29</u> | | | 98.40 | |
| DSC-L2 [17], reported from [19] | 93.68 | 94.08 | - | - | 93.97 | - |
| DASC [21] | 96.39 | 96.86 | - | - | 96.32 | - |
| MESC [18] | 98.40 | 98.29 | - | - | 98.40 | - |
| SAE [23] | 86.29 | 90.28 | - | - | - | - |
| DCSC [24] | 92.08 | 95.39 | - | - | - | - |
| LBDR [25] | 78.59 | 86.97 | - | - | - | - |

TABLE IX: Clustering results on COIL100 dataset. Sub-bands for each WPD SC algorithm are reported in table.

| Algorithm | ACC [%] in-sample out-of-sample | NMI [%] in-sample out-of-sample | Rand [%] in-sample out-of-sample | F_score[%] in-sample out-of-sample | Purity [%] in-sample out-of-sample | Average affinity |
|---|---|---|---|---|---|---|
| WP MERA [27] | 84.59±1.85 80.01±1.80 | 94.25±0.51 90.73±0.61 | 80.40±1.83 72.80±1.89 | 80.60±1.81 73.07±1.87 | 88.61±1.47 81.86±1.52 | |
| IPD, d=10 | **87.45±1.49** **82.39±1.62** | **94.73±0.59** **91.18±0.64** | **82.82±1.98** **74.63±1.84** | **82.99±1.96** **74.88±1.82** | **88.89±1.48** **83.65±1.41** | |
| SSC [9] | 51.17±1.33 51.26±1.25 | 78.08±0.69 77.80±0.69 | 41.16±1.78 41.12±1.77 | 41.84±1.75 41.80±1.74 | 58.50±0.97 57.93±0.98 | 0.3676±0.0025 |
| IPD, d=2 | 64.57±1.87 61.64±1.68 | 85.83±0.66 82.14±0.66 | 55.39±2.35 51.91±2.06 | 55.89±2.31 52.45±2.03 | 69.07±1.59 66.14±1.37 | 0.6250±0.0030 |
| WP-AH SSC | 62.36±1.51 61.50±1.46 | 84.52±0.80 83.56±0.77 | 49.13±4.13 48.54±3.88 | 49.75±4.05 49.17±3.81 | 67.80±1.13 66.81±1.14 | 0.2975±0.0028 |
| IPD, d=3 | 67.24±1.57 65.20±1.50 | 87.64±0.63 84.68±0.67 | 55.26±3.67 53.64±2.47 | 55.80±3.60 54.19±2.42 | 71.82±1.30 69.64±1.21 | 0.2544±0.0030 |
| GMC LRSSC [10] | 47.95±1.28 47.82±1.25 | 74.26±0.56 74.53±0.53 | 38.07±1.35 38.40±1.31 | 38.75±1.33 39.08±1.28 | 53.26±0.99 52.71±1.06 | 0.3765±0.0032 |
| WP-AH GMC LRSSC | 53.82±1.44 53.62±1.28 | 77.95±0.71 77.65±0.62 | 37.97±2.63 41.11±1.95 | 38.74±2.57 41.80±1.91 | 59.72±1.18 59.26±1.05 | 0.3008±0.0032 |
| S0L0 LRSSC [10] | 50.47±1.19 49.86±1.18 | 75.52±0.40 75.44±0.40 | 43.51±1.01 43.13±1.06 | 44.09±1.00 43.72±1.04 | 53.64±0.90 52.87±0.91 | 0.3828±0.0022 |
| WP-AH S0L0 LRSSC | 54.96±1.39 54.32±1.32 | 79.53±0.59 79.09±0.56 | 46.62±1.65 47.01±1.47 | 47.23±1.62 47.60±1.45 | 60.51±1.01 59.58±1.03 | 0.3037±0.0025 |
| LRR [8] | 36.84±2.30 38.77±2.08 | 69.20±1.83 72.18±1.40 | 15.41±2.49 20.04±4.32 | 16.79±4.15 21.26±4.19 | 43.19±1.96 44.60±1.84 | 0.3525±0.0033 |
| IPD, d=9 | 56.73±1.47 | 82.67±0.48 | 45.95±2.47 | 46.62±2.42 | 61.63±1.06 | 0.3802±0.0069 |
| | 57.29±1.46 | 83.13±0.52 | 47.24±1.99 | 47.86±1.96 | 61.98±1.22 | |
| WP-AH LRR | 34.92±1.90 | 68.30±1.33 | 11.89±2.39 | 13.42±2.31 | 43.51±1.55 | 0.2751±0.0033 |
| | 37.80±2.01 | 71.48±1.20 | 15.59±2.19 | 16.97±2.13 | 46.31±1.79 | |
| IPD, d=4 | 58.24±2.14 | 83.31±0.83 | 33.21±4.94 | 34.22±4.82 | 64.10±1.66 | 0.2553±0.0039 |
| | 58.53±2.03 | 81.59±0.73 | 40.71±2.48 | 41.48±2.42 | 64.69±1.56 | |
| NSN [67] | 57.15±1.11 | 79.88±0.43 | 49.20±1.06 | 49.73±1.05 | 60.21±1.03 | 0.3718±0.0019 |
| | 57.17±0.98 | 80.23±0.46 | 48.28±1.08 | 48.81±1.07 | 60.24±0.92 | |
| WP-AA NSN | 79.64±1.42 | 91.32±0.40 | 74.28±1.34 | 74.54±1.33 | 81.12±1.21 | 0.5132±0.0013 |
| | 78.25±1.21 | 90.63±0.46 | 71.93±1.30 | 72.21±1.28 | 79.74±1.10 | |
| RTSC [11] | 59.87±1.63 | 84.37±0.41 | 49.98±1.73 | 50.58±1.70 | 66.36±1.12 | 0.3624±0.0044 |
| | 60.33±1.72 | 84.54±0.50 | 49.33±1.70 | 49.91±1.67 | 66.42±1.23 | |
| WP-AA RTSC | 68.78±1.07 | 86.81±0.27 | 62.04±1.10 | 62.45±1.08 | 72.90±0.91 | 0.5040±0.0028 |
| | 68.61±1.27 | 86.77±0.39 | 60.85±1.35 | 61.26±1.33 | 72.55±0.87 | |
| $S_{1/2}$-LRR [68] | 48.62±1.29 | 74.09±0.56 | 41.62±1.29 | 42.23±1.28 | 52.38±1.02 | 0.3852±0.0028 |
| | 49.94±1.12 | 75.72±0.48 | 41.54±1.35 | 42.14±1.34 | 53.51±1.02 | |
| IPD, d=4 | 56.89±1.34 | 82.22±0.47 | 48.58±1.64 | 49.17±1.61 | 62.47±0.97 | 0.4836±0.0027 |
| | 57.34±1.53 | 82.34±0.59 | 48.10±1.91 | 48.68±1.88 | 62.53±1.27 | |
| WP-AH $S_{1/2}$-LRR | 53.43±1.21 | 75.84±0.57 | 45.55±1.30 | 46.11±1.28 | 57.32±0.99 | 0.3255±0.0020 |
| | 54.56±1.24 | 77.33±0.60 | 45.66±1.49 | 46.21±1.47 | 58.05±1.98 | |
| IPD, d=4 | 52.34±1.13 | 76.76±0.47 | 42.70±1.19 | 43.36±1.17 | 57.91±0.84 | 0.2708±0.0018 |
| | 53.30±1.13 | 78.11±0.51 | 43.67±1.16 | 44.29±1.15 | 58.58±0.95 | |
| $S_{2/3}$-LRR [68] | 50.00±1.22 | 74.92±0.44 | 43.31±0.99 | 43.89±0.97 | 53.21±1.03 | 0.3824±0.0023 |
| | 51.01±1.33 | 76.21±0.46 | 42.78±1.05 | 43.36±1.04 | 54.07±1.09 | |
| IPD, d=4 | 54.92±1.05 | 80.81±0.49 | 47.57±1.37 | 48.17±1.35 | 60.47±0.89 | 0.4866±0.0022 |
| | 54.96±1.10 | 81.19±0.47 | 46.87±1.40 | 47.46±1.38 | 60.23±0.98 | |
| WP-AH $S_{2/3}$-LRR | 53.79±1.23 | 76.19±0.58 | 45.98±1.32 | 46.54±1.31 | 57.78±1.09 | 0.3244±0.0018 |
| | 55.30±1.29 | 77.77±0.60 | 46.47±1.46 | 47.00±1.44 | 58.79±1.18 | |
| IPD, d=4 | 49.37±1.09 | 74.44±0.45 | 39.01±0.97 | 39.72±0.96 | 55.02±0.94 | 0.2665±0.0027 |
| | 50.61±1.02 | 76.53±0.39 | 40.55±0.93 | 41.22±0.91 | 56.04±0.72 | |
| **Deep networks** | | | | | | |
| MAESC [18] | 71.88 | 90.76 | | | | |

| DSLSP [70], reported from [18] | 65.86 | 89.14 | - | - | - | - |
|---|---|---|---|---|---|---|
| DSC-L2 [17], reported from [18] | 67.71 | 89.08 | - | - | - | - |
| LRAE [70], reported from [18] | 56.62 | 79.77 | | | | |
| DSCNS S [71] | 71.42 | | | | | |
| DSRSCN [72] | 72.53 | 72.94 | | | | |
| DCFSC [73] | 72.70 | | | | | |
| SAE [23] | 55.80 | 76.12 | - | - | - | - |
| DCSC [24] | 60.27 | 82.26 | - | - | - | - |